\PassOptionsToPackage{table}{xcolor}
\documentclass[letterpaper]{article} 
\usepackage[submission]{aaai2026}  
\usepackage{times}  
\usepackage{helvet}  
\usepackage{courier}  
\usepackage[hyphens]{url}  
\usepackage{graphicx} 
\usepackage{natbib}  
\usepackage{caption} 
\usepackage{algorithm}
\usepackage{algorithmic}

\usepackage{booktabs}
\usepackage{xcolor}
\usepackage{amsmath}
\usepackage{multirow}
\usepackage{pifont}
\usepackage{newfloat}
\usepackage{listings}
\DeclareCaptionStyle{ruled}{labelfont=normalfont,labelsep=colon,strut=off} 
\floatstyle{ruled}
\newfloat{listing}{tb}{lst}{}
\floatname{listing}{Listing}
\title{F$^2$RVLM: Boosting Fine-grained Fragment Retrieval \\ for Multi-Modal Long-form Dialogue with Vision Language Model}
\author{
    Hanbo Bi\equalcontrib, Zhiqiang Yuan\equalcontrib, Zexi Jia, Jiapei Zhang, Chongyang Li, \\
    Peixiang Luo, Ying Deng, Xiaoyue Duan, Jinchao Zhang\thanks{Corresponding author: Jinchao Zhang}
}
\affiliations{
    Pattern Recognition Center, WeChat AI, Tencent Inc, China\\

}

\usepackage{bibentry}

\begin{document}

\maketitle

\begin{abstract}
Traditional dialogue retrieval aims to select the most appropriate utterance or image from recent dialogue history. However, they often fail to meet users’ actual needs for revisiting semantically coherent content scattered across long-form conversations.
To fill this gap, we define the Fine-grained Fragment Retrieval (FFR) task, requiring models to locate query-relevant fragments, comprising both utterances and images, from multimodal long-form dialogues.
As a foundation for FFR, we construct MLDR, the longest-turn multimodal dialogue retrieval dataset to date, averaging 25.45 turns per dialogue, with each naturally spanning three distinct topics. To evaluate generalization in real-world scenarios, we curate and annotate a WeChat-based test set comprising real-world multimodal dialogues with an average of 75.38 turns.
Building on these resources, we explore existing generation-based Vision-Language Models (VLMs) on FFR and observe that they often retrieve incoherent utterance-image fragments. While optimized for generating responses from visual-textual inputs, these models lack explicit supervision to ensure semantic coherence within retrieved fragments.
To address this, we propose F$^2$RVLM, a generative retrieval model trained in a two-stage paradigm: (1) supervised fine-tuning to inject fragment-level retrieval knowledge, and (2) GRPO-based reinforcement learning with multi-objective rewards to encourage outputs with semantic precision, relevance, and contextual coherence.
In addition, to account for difficulty variations arising from differences in intra-fragment element distribution, ranging from locally dense to sparsely scattered, we introduce a difficulty-aware curriculum sampling that ranks training instances by predicted difficulty and gradually incorporates harder examples. This strategy enhances the model’s reasoning ability in long, multi-turn dialogue contexts.
Experiments on both in-domain and real-domain sets demonstrate that F$^2$RVLM substantially outperforms popular VLMs, achieving superior retrieval performance. Code and dataset are available at  \textcolor{blue}{\url{https://f2rvlm.github.io}}.

\end{abstract}


\section{Introduction}

\begin{figure}[t]
\centering
\includegraphics[width=1.0\linewidth]{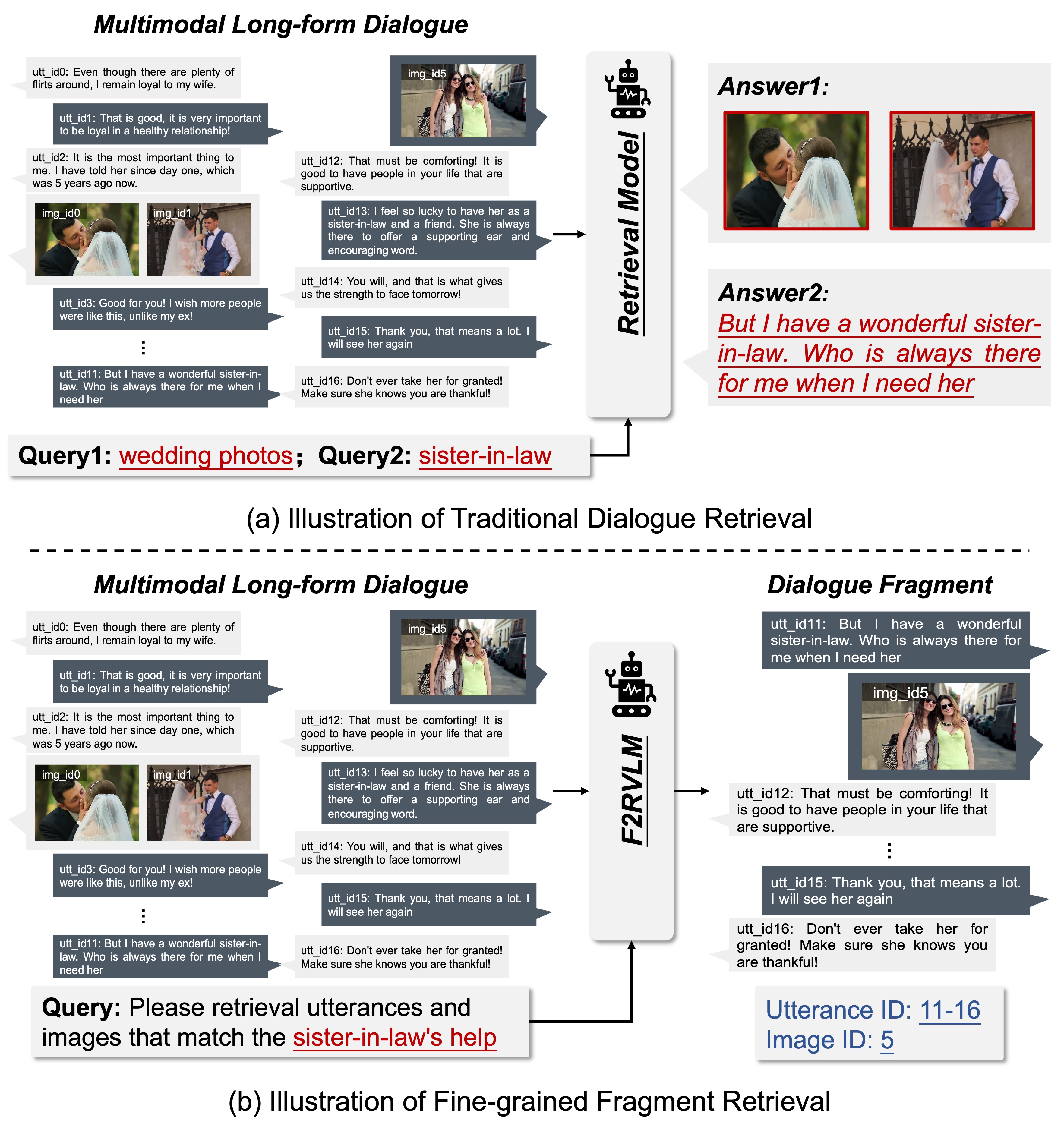}
\vspace{-15pt}  
\caption{Comparison between the traditional Dialogue Retrieval and our Fine-grained Fragment Retrieval task.}
\vspace{-15pt}  
\label{fig:contrast}
\end{figure}

With the widespread adoption of messaging platforms and AI assistants, users frequently need to revisit earlier utterances or referenced images for information confirmation or context tracking. Efficiently retrieving specific content from ongoing multimodal dialogues has become a crucial capability for intelligent systems. Such dialogues often span dozens of turns, intertwining multiple topics, user intents, and visual-textual cues. As illustrated in Fig.\ref{fig:contrast}, unlike conventional retrieval tasks that select the most appropriate element from the dialogue history, this scenario demands models to directly locate semantically coherent fragments, including both text and images, enabling users to efficiently access relevant historical content. To address this need, we define the \textbf{Fine-grained Fragment Retrieval (FFR)} task, which aims to accurately extract query-relevant fragments from long-form, multi-party, multimodal dialogues.

As a foundation for studying FFR, we construct MLDR, a large-scale multimodal long-form dialogue retrieval dataset. Each dialogue covers three distinct topics with an average of 25.45 turns, making it the longest-turn multimodal dialogue dataset to date. To support diverse retrieval intents, we provide multi-granularity annotations, including domain-level and event-level tags for each dialogue unit.
In addition, we curate a WeChat-based test set comprising real-world multimodal dialogues with an average of 75.38 turns, aiming to evaluate generalization in practical settings. This set is collected from real conversations contributed by multiple volunteers and annotated through a multi-stage process, serving as a realistic and challenging benchmark for fragment retrieval (see Appendix.\textcolor{red}{A} for details).

Building on our constructed dataset resources, we investigate existing retrieval models on the proposed FFR task. In retrieval scenarios, current models predominantly follow two paradigms: embedding-based and generation-based. Embedding-based Vision-Language Models (VLMs), such as CLIP~\cite{radford2021learning} and BLIP2~\cite{li2023blip}, encode images and texts into a shared embedding space and retrieve via similarity ranking. In contrast, generation-based models are typically pre-trained on large-scale image-text corpora to align cross-modal semantics, and subsequently instruction-tuned to generate responses conditioned on visual-textual inputs. Models like GPT-4o~\cite{jaech2024openai} and Gemini~\cite{comanici2025gemini} demonstrate strong performance in prompt-driven retrieval settings, leveraging unified architectures with promising contextual understanding in open-ended multimodal dialogues.

Despite their general success, our evaluation reveals that even leading VLMs struggle to localize relevant fragments accurately in long-form multimodal dialogues. Models such as Qwen2.5-VL-72B~\cite{wang2024qwen2} and Doubao-Seed-1.6~\cite{guo2025seed1} frequently retrieve incoherent utterance-image pairs, e.g., mismatched dialogue turns or irrelevant visual content, resulting in suboptimal F1 scores under real-world conditions (see qualitative examples in Appendix.\textcolor{red}{C}). This limitation primarily stems from the gap between the models’ learning objectives and the demands of fragment retrieval: while optimized for generating responses from visual-textual inputs, these models lack explicit supervision to ensure that the retrieved or generated fragments (both utterances and images) are semantically coherent and contextually aligned with the user query. 

To address these limitations, we propose \textbf{F$^2$RVLM}, a generation-based retrieval framework for long-form multimodal dialogues, tailored for \textbf{F}ine-grained \textbf{F}ragment \textbf{R}etrieval with \textbf{V}ision-\textbf{L}anguage \textbf{M}odel. F$^2$RVLM follows a two-stage training paradigm: fragment-level retrieval knowledge is first injected via supervised fine-tuning, followed by GRPO-based reinforcement learning to align retrieval behavior with human preferences.
We design a multi-objective reward scheme that encourages the generation of fragments with semantic precision and contextual coherence. Specifically: (1) an F1-based alignment reward encourages accurate matching with ground-truth fragments, penalizing both over- and under-retrieval; (2) a fragment order consistency enhances semantic alignment between selected utterances and images, guiding the model to organize content in a coherent, human-preferred manner. 
Moreover, we observe that fragment structures in long-form dialogues differ in their internal distribution, from locally dense to sparsely scattered, which directly reflects differences in retrieval difficulty. To exploit this inherent hierarchy of difficulty, we introduce a curriculum sampling that ranks training samples by predicted F1 and confidence, progressively exposing the model to harder instances with greater contextual complexity. This promotes robust reasoning in diverse, noisy, and long-range scenarios.
Extensive experiments on both the in-domain MLDR and real-domain WeChat-based sets demonstrate that F$^2$RVLM significantly outperforms mainstream VLMs in retrieval accuracy and contextual understanding. 

Our main contributions are summarized as follows:


\begin{itemize}
\item We introduce Fine-grained Fragment Retrieval, a novel retrieval task that aims to directly locate semantically coherent utterance-image fragments from long-form dialogues, differing from traditional dialogue retrieval that selects the most appropriate individual elements.
\item  We construct MLDR, the longest-turn multimodal dialogue retrieval dataset to date, averaging 25.45 turns per dialogue, with each covering three distinct topics. Additionally, we curate a real-world WeChat-based test set averaging 75.38 turns per dialogue to evaluate retrieval generalization in practice.

\item We propose F$^2$RVLM, a generation-based retrieval model for the proposed task. It integrates GRPO-based reinforcement learning with multi-objective rewards and difficulty-aware curriculum sampling to progressively enhance the semantic consistency and completeness of retrieved fragments.

\item Experimental results on both the in-domain MLDR validation set and the real-domain WeChat-based test set demonstrate that F$^2$RVLM consistently outperforms popular VLMs in fragment retrieval accuracy.

\end{itemize}
    
\section{Related Work}
\textbf{Multimodal Dialogue Datasets.}
Recent advances in vision-language modeling have accelerated progress in multimodal dialogue understanding~\cite{meng2020openvidial}. Existing dialogue datasets fall into two main types: (1) Image-grounded datasets~\cite{shuster2020image, zheng2022mmchat, lin2023tiktalk} consist of dialogues explicitly constructed around a given image, often collected via crowdsourcing. While well-aligned, they lack the natural heterogeneity of real conversations, where not all utterances refer to images. (2) Image-sharing datasets\cite{zang2021photochat, lee2021constructing, feng2023mmdialog, lee2024dialogcc} address this by capturing more spontaneous visual usage. For instance, MMDialog~\cite{feng2023mmdialog} collects over 1M social media conversations with images dispersed across turns, while DialogCC~\cite{lee2024dialogcc} enriches textual dialogues by inserting suitable images. However, these corpora still consist mostly of short, single-topic dialogues and lack the long-range, multi-topic structure of real-world interactions.

\begin{figure*}[t]
\centering
\includegraphics[width=0.92\linewidth]{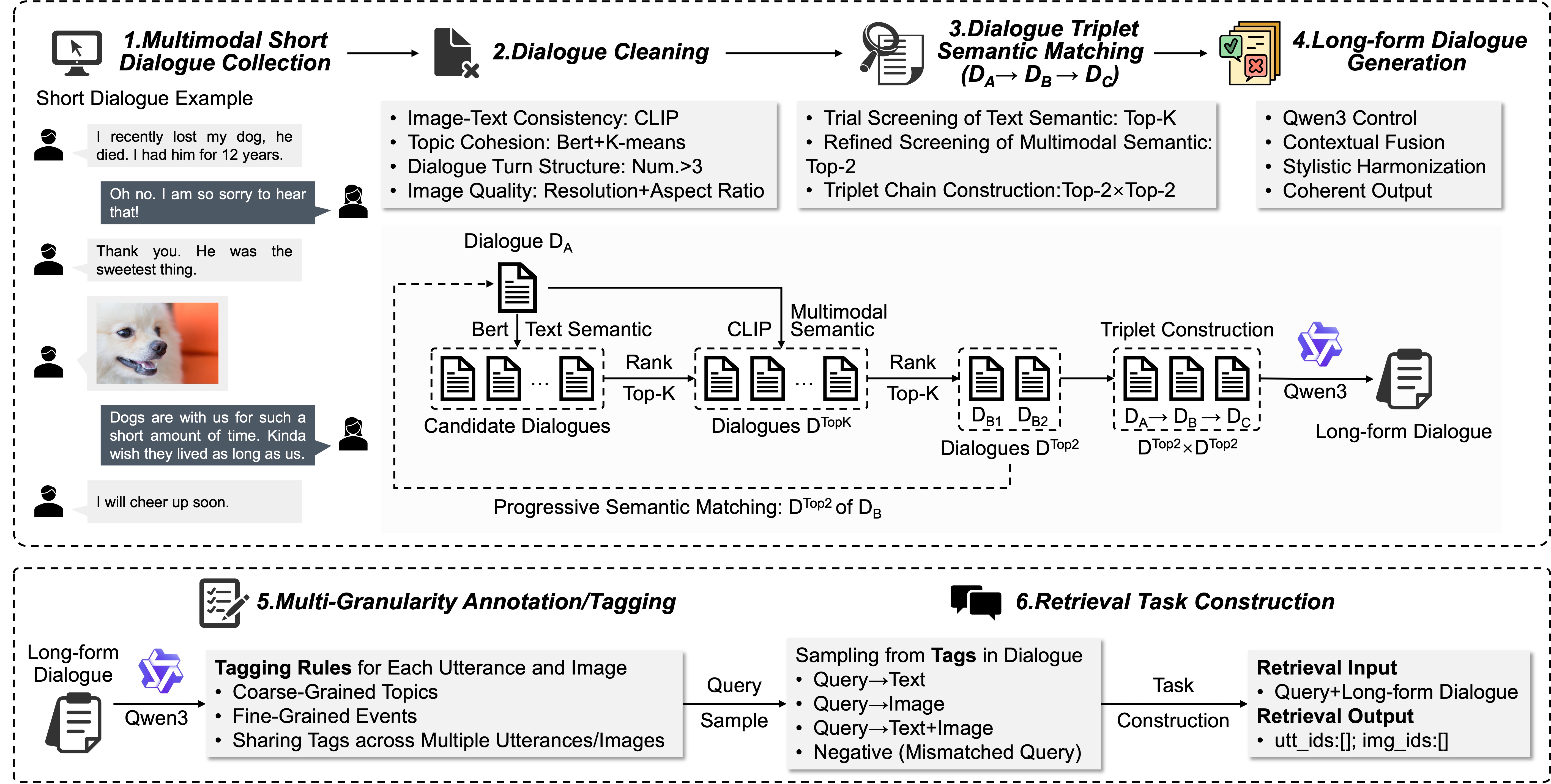}
\vspace{-3pt}  
\caption{Overview of the MLDR construction pipeline, which integrates multimodal short dialogue processing, Qwen3-driven long-form dialogue generation, multi-granularity annotation, and retrieval-oriented task design.}
\vspace{-15pt}  
\label{fig:1}
\end{figure*}

\noindent \textbf{VLMs for Dialogue Retrieval.} 
VLMs have achieved notable success in image-text retrieval, typically following two paradigms: (1) embedding-based VLMs~\cite{radford2021learning, li2022blip, liu2022universal, zhang2024gme, wei2024uniir} that learn joint visual-text representations for efficient matching, and (2) generation-based VLMs~\cite{liu2023visual, achiam2023gpt, wang2024qwen2, jiang2024vlm2vec, liu2025lamra} that combine LLMs with visual encoders to generate context-aware responses via multimodal prompts. Recently, multimodal dialogue retrieval has gained traction, focusing on selecting the most appropriate sentence or image given the dialogue history~\cite{yin2024dialclip, choe2025multimodal}. For example, DRIBER~\cite{lee2023large} and VCU~\cite{wei2024balancing} generate intermediate image descriptions for dialogue-to-image retrieval, while IGSR~\cite{wang2025new} retrieves stickers by identifying user intent. However, these tasks are constrained to short-context response selection. In this work, we investigate a more challenging yet underexplored setting, Fine-grained Fragment Retrieval, which aims to retrieve semantically aligned fragments scattered across long-form, multi-topic dialogues.

\noindent \textbf{VLMs with Reinforcement Learning.}
Recent advances like OpenAI’s o1~\cite{jaech2024openai} and DeepSeek-R1~\cite{guo2025deepseek} have shifted LLM research toward enhancing reasoning via reinforcement learning (RL), leading to paradigms such as GRPO~\cite{shao2024deepseekmath}, DAPO~\cite{yu2025dapo}, and VAPO~\cite{yue2025vapo}. This trend is expanding into the multimodal domain~\cite{shen2025vlm,wang2025vl}, with methods like LMM-R1~\cite{peng2025lmm} introducing rule-based two-stage RL, and Reason-RFT~\cite{tan2025reason} using SFT with CoT data for RL initialization. Vision-R1~\cite{huang2025vision} employs progressive reflection suppression in GRPO, while Visual-RFT~\cite{liu2025visual} enhances visual reasoning via verifiable rewards. Video-R1~\cite{feng2025video} further extends GRPO with temporal modeling for video-language tasks.
In this work, we explore RL for multimodal dialogue content retrieval, aiming to improve fine-grained reasoning and fragment localization in long-form image-text contexts.


\section{MLDR Dataset}
This section outlines the construction of MLDR and the WeChat test set, along with their statistical analysis.

\subsection{MLDR Construction}
As highlighted in Table~\ref{tab:1}, existing multimodal dialogue datasets primarily consist of short, single-topic conversations, limiting their utility in modeling the multi-topic nature of real-world dialogues. To fill this gap, we construct a multimodal long-form dialogue corpus and further develop the MLDR dataset. The pipeline is illustrated in Fig.\ref{fig:1} and summarized below (see Appendix.\textcolor{red}{A} for details):

\begin{table}[t]
\centering
\resizebox{0.85\linewidth}{!}{
\begin{tabular}{cccc} 
\toprule
\textbf{\#}    & \textbf{Datasets}                                                      & \textbf{Avg. Turn}                                 & \textbf{Avg. Topic}                                \\ 
\hline
\multirow{6}{*}{ENG} & ImageChat~\cite{shuster2020image}                                                                                                    & 1.98~                                              & 1.00~                                              \\
& OpenViDial~\cite{meng2020openvidial}                                                                                      & 1.00~                                              & 1.00~                                              \\
& PhotoChat~\cite{zang2021photochat}                                                                                          & 12.74~                                             & 1.00~                                              \\
& MMDD~\cite{lee2021constructing}                                                                                                 & 11.56~                                             & 1.00~                                              \\
& MMDialog~\cite{feng2023mmdialog}                                                                                            & 4.56~                                              & 1.00~                                              \\
& DialogCC~\cite{lee2024dialogcc}                                                                                        & 8.20~                                              & 1.00~                                              \\ 
\hline
\multirow{6}{*}{CN} & MMChat~\cite{zheng2022mmchat}                                                                                               & 2.59~                                              & 1.00~                                              \\
& M3ED~\cite{zhao2022m3ed}                                                                                                       & 9.17~                                              & 1.00~                                              \\
& CPED~\cite{chen2022cped}                                                                                                 & 11.08~                                             & 1.00~                                              \\
& CMMA~\cite{zhang2023cmma}                                                                                                  & 7.27~                                              & 1.00~                                              \\
& TikTalk~\cite{lin2023tiktalk}                                                                                               & 2.25~                                              & 1.00~                                              \\
& {\cellcolor[rgb]{0.951,0.951,0.951}}\textbf{\textbf{Our MLDR Dataset}} & {\cellcolor[rgb]{0.951,0.951,0.951}}\textbf{25.45} & {\cellcolor[rgb]{0.951,0.951,0.951}}\textbf{3.00}  \\
\bottomrule
\end{tabular}
}
\vspace{-3pt}  
\caption{Summary of main multimodal dialogue datasets.} 
\label{tab:1}
\vspace{-15pt}
\end{table}

\noindent \textbf{Short Dialogue Collection \& Clearning.} 
We begin with topic-rich short dialogues from DialogCC, providing a semantically diverse foundation for coherent long-form generation. To ensure data quality, we apply a multi-criteria cleaning pipeline to remove samples with poor image-text alignment, topic drift, unstructured turns, or low image quality.

\noindent \textbf{Dialogue Triplet Semantic Matching.}
To construct coherent long-form dialogues, we design a triplet-based matching strategy that ensures topic continuity and multimodal alignment. For each short dialogue $D_A$, we first identify the Top-K semantically similar candidates using BERT-based sentence embeddings. We then refine this set by computing a weighted multimodal similarity score with CLIP (0.7 for text, 0.3 for image), and select the Top-2 most aligned dialogues $D_B$. This process is repeated for each $D_B$ to retrieve two additional candidates $D_C$, forming four unique triplets of the form $D_A \rightarrow D_B \rightarrow D_C$, which serve as structural units for long-form synthesis.

\noindent \textbf{Long-form Dialogue Generation.} 
Each matched triplet is converted into a coherent long-form dialogue utilizing the Qwen3-256B~\cite{yang2025qwen3} model with structured prompts. The generation process preserves multimodal semantics, ensures topical coherence through smooth transitions, and yields fluent, contextually grounded long-form dialogues, forming high-quality samples for downstream retrieval and reasoning tasks.



\noindent \textbf{Multi-Granularity Annotation.} 
To support fine-grained fragment retrieval, we adopt a two-level shared tagging scheme automatically generated by the Qwen3 under structured prompts. Each sentence and image caption is assigned a coarse-grained tag (e.g., domain) and a fine-grained tag (e.g., event), with each tag shared by at least two elements to form semantically consistent fragments. 


\noindent \textbf{Task Construction for Retrieval.} 
We formulate a unified retrieval task over annotated long-form dialogues, where sampled coarse- or fine-grained tags serve as natural language queries, and corresponding utterances/images sharing the same tag form the retrieval fragments. To enhance diversity and robustness, we adopt a query-driven sampling strategy that generates four types of query-fragment pairs: (a) multimodal fragments; (b) utterance-only fragments; (c) image-only fragments; and (d) negative samples constructed by replacing queries with unrelated tags. This formulation supports generalization across varied retrieval scenarios.



\subsection{Real-domain Evaluation on WeChat Dialogues}
To evaluate fine-grained retrieval in real-world scenarios, we construct a real-domain test set from naturally occurring multimodal WeChat conversations. Unlike the synthesized MLDR data, these dialogues reflect authentic user interactions, with noisy inputs, informal expressions, and frequent topic shifts, posing greater challenges for robust retrieval.

We collect multi-turn image-utterance chats from 12 volunteers\footnotemark\footnotetext{Volunteers manually provided their own chat records from the application. We emphasize that all data were solely provided by volunteers, and no chat records were, or will ever be, obtained from WeChat's backend.} and preprocess them by removing emojis, sensitive content, profanity, and semantically void utterances. Consecutive messages from the same speaker are merged, and only dialogues with at least two images are retained. This yields 270 long-form dialogues (avg. 145.38 turns), which are segmented if exceeding 100 turns, resulting in 580 coherent samples. All dialogues are manually annotated by a professionally trained team, producing 1,250 query-dialogue pairs. Each sample includes a natural language query, its multimodal dialogue (avg. 75.38 turns), and ground-truth utterance and image IDs as retrieval targets. The task format remains consistent with MLDR, enabling direct evaluation of model generalization in open-domain fragment retrieval.

\begin{figure}[t]
\centering
\includegraphics[width=1.0\linewidth]{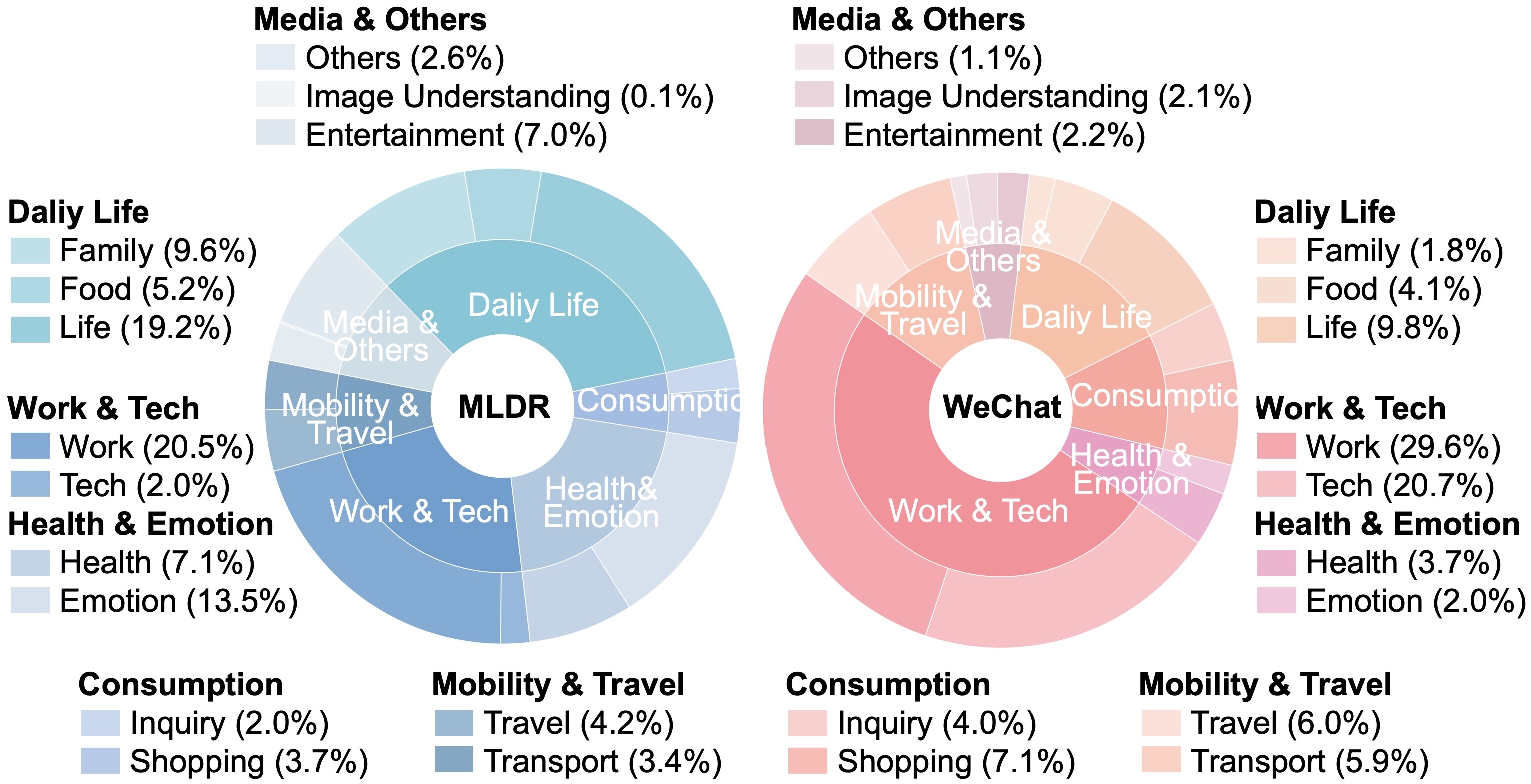}
\vspace{-15pt}  
\caption{Comparison between the MLDR and WeChat-based datasets in terms of dialogue topic distributions.}
\vspace{-18pt}  
\label{fig:2}
\end{figure}

\begin{figure*}[t]
\centering
\includegraphics[width=0.94\linewidth]{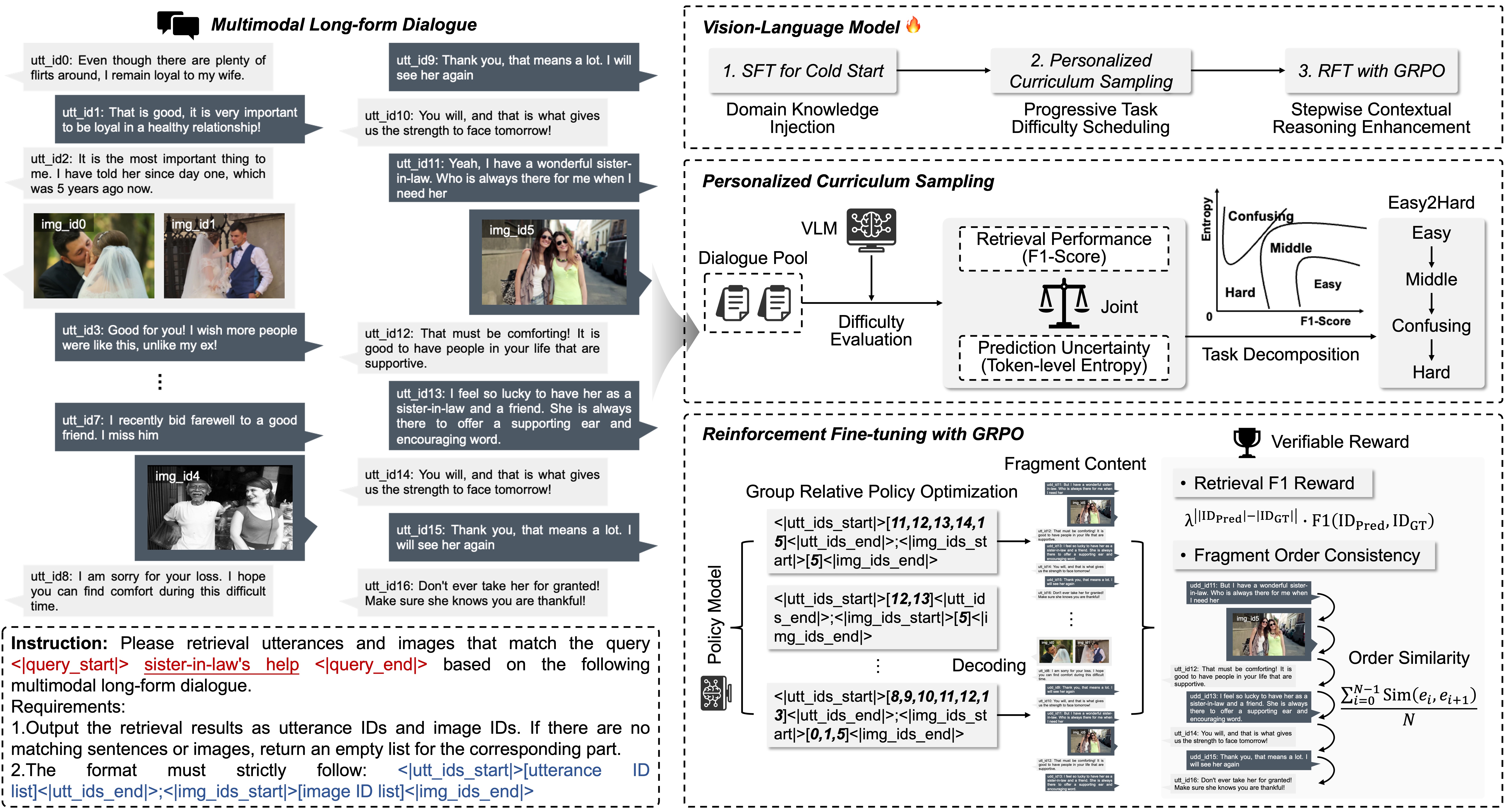}
\vspace{-3pt}  
\caption{Overview of the F$^2$RVLM framework for multimodal long-form dialogue fragment retrieval. It consists of supervised fine-tuning for fragment-level knowledge injection, personalized curriculum sampling based on retrieval difficulty, and GRPO-based reinforcement learning to jointly enhance retrieval accuracy and fragment semantic coherence.}
\label{fig:3}
\vspace{-15pt}  
\end{figure*}

\subsection{Dataset Statistics and Analysis}
To better understand the characteristics of MLDR and its real-world generalization benchmark, we compare topic diversity between MLDR and the WeChat-based test set (Fig.\ref{fig:2}). We organize dialogue content into six high-level domains (e.g., Daily Life, Work \& Tech), further divided into 14 fine-grained topics. Each MLDR dialogue is deliberately constructed to span three distinct topics, promoting multi-topic contextual modeling and ensuring a balanced semantic distribution. In contrast, the WeChat-based test set exhibits natural domain bias, with over 50\% of conversations centered on Work \& Tech. Further comparisons of dialogue turn distributions are provided in the Appendix.\textcolor{red}{A}.


\section{Methodology}

\subsection{Task Definition} \label{sec:task_definition}
Fine-grained fragment retrieval (FFR) can be formulated as a structured prediction task over long-form, multi-turn dialogues that include both textual utterances and images. Given a multimodal dialogue $D = \{(u_1, m_1), (u_2, m_2), \dots, (u_T, m_T)\}$, where each turn $t$ consists of a speaker $u_t$ and a message $m_t$ that may contain text or images, and a user-issued query $q$, the objective is to retrieve a subset of utterance and image IDs that are semantically relevant to the query. To enable the VLMs to perceive and understand the semantic and temporal structure of each turn, explicit structural markers are inserted into the dialogue: \texttt{<|utt\_id\_start|>}...\texttt{<|utt\_id\_end|>} to mark each utterance with a unique ID, while \texttt{<|img\_id\_start|>}...\texttt{<|img\_id\_end|>} marks each embedded image. In summary, the retrieval process can be formalized as:
\begin{align} \label{equation:2}
\hat{y} = \mathcal{F}(D, q) = (\hat{I}_{\text{utt}}, \hat{I}_{\text{img}})
\end{align}
where $\hat{I}_{\text{utt}}$ and $\hat{I}_{\text{img}}$ denote the predicted sets of relevant utterance and image IDs, respectively.

\subsection{Framework Overview}

As illustrated in Fig.\ref{fig:3}, given a long-form dialogue $D$ and a query $q$, F$^2$RVLM generates structured outputs of relevant utterance and image IDs. To enable this behavior, the model is trained in two stages: first, supervised fine-tuning injects task-specific knowledge for fragment retrieval; then, GRPO-based reinforcement fine-tuning is conducted with a multi-objective reward that promotes semantic precision and contextual coherence. To stabilize training and enhance long-range reasoning, samples are organized into a difficulty-aware curriculum based on predicted F1 and uncertainty.

\subsection{Optimize Fragment Semantic Coherence by GRPO}
We adopt a rule-based variant of Group-Relative Policy Optimization (GRPO)~\cite{shao2024deepseekmath} to align model behavior with human-preferred retrieval patterns. To this end, we design three verifiable reward functions targeting format compliance, retrieval accuracy, and fragment order consistency, jointly guiding the model to generate structured, precise, and contextually coherent outputs.


\noindent \textbf{Format Reward} $(R_{\text{Format}})$. 
$R_{\text{Format}}$ encourages the model to strictly adhere to the expected output format:
\begin{itemize}
    \item \texttt{<|utt\_ids\_start|>[...]<|utt\_ids\_end|>};
    \item \texttt{<|img\_ids\_start|>[...]<|img\_ids\_end|>}
\end{itemize}
This binary reward returns 1 only when the output fully matches the required format. A uniqueness constraint is also enforced: if any duplicate IDs are present, the reward is set to 0, even if the overall format appears correct.


\noindent \textbf{Retrieval F1 Reward} $(R_{\text{F1}})$. 
To encourage precise yet well-scoped retrieval, we design a reward function based on the F1 score with an exponential penalty on length deviation. It considers both utterance and image ID overlaps between predictions and ground truth, while explicitly penalizing over- and under-retrieval:
\begin{equation}
R_{\text{F1}} =  \sum_{m \in \{\text{utt}, \text{img}\}} \lambda_m \cdot \mathrm{F1}(I_m^{\text{pred}}, I_m^{\text{gt}}) \cdot \gamma^{|\,|I_m^{\text{pred}}| - |I_m^{\text{gt}}|\,|}
\end{equation}
where $I_{\text{*}}^{\text{pred}}$ and $I_{\text{*}}^{\text{gt}}$ denote the predicted and ground-truth ID sets, respectively. The weighting factors $\lambda$ balances modality importance. The penalty base $\gamma \in (0,1)$ modulates the severity of the length penalty, which increases exponentially with deviation in prediction length. 
\begin{table*}[t]
\centering
\resizebox{0.86\linewidth}{!}{
\begin{tabular}{r|cccc|cccc} 
\toprule
\multirow{2}{*}{Model}           & \multicolumn{4}{c}{In-domain Val Set (MLDR)}                      & \multicolumn{4}{c}{Real-domain Test Set (WeChat)}                  \\ 
\cline{2-9}
                                 & Precision(\%)  & Recall(\%)     & F1(\%)         & MCC(\%)        & Precision(\%)  & Recall(\%)     & F1(\%)         & MCC(\%)         \\ 
\hline
CLIP-Embedding$^\dagger$                    & 48.74          & 30.80          & 42.73          & 21.14          & 20.44          & 49.91          & 31.56          & 17.62           \\
BLIP2-Embedding$^\dagger$                  & 31.88          & 2.96           & 23.47          & 0.00           & 15.22          & 52.74          & 25.05          & 4.94            \\
E5-V-Embedding$^\dagger$                   & 53.85          & 48.83          & 51.82          & 28.92          & 30.33          & 47.13          & 36.99          & 24.40           \\
GME-Embedding$^\dagger$                    & 62.27          & 23.75          & 35.52          & 24.17          & 29.63          & 53.11          & 38.22          & 26.68           \\
Qwen2.5-VL-7B$^\dagger$                    & 21.20          & 4.27           & 7.84           & 0.00           & 10.22          & 16.34          & 12.58          & 0.00            \\
MiMo-7B-RL$^\dagger$                       & 67.68          & 55.19          & 61.30          & 45.52          & 41.74          & 24.48          & 30.94          & 23.68           \\
Qwen2.5-VL-72B$^\dagger$                   & 61.11          & 67.14          & 64.09          & 44.61          & 32.55          & 36.95          & 36.60          & 25.26           \\
Doubao-Seed-1.6$^\dagger$                  & 73.83          & 42.19          & 54.67          & 42.57          & 55.22          & 28.47          & 38.63          & 34.06           \\
Claude-Sonnet-4$^\dagger$                  & 67.21          & 58.09          & 62.80          & 46.57          & 51.15          & 40.88          & 46.30          & 40.50           \\
GPT-4o$^\dagger$                           & 70.43          & 52.49          & 60.32          & 45.91          & \underline{56.11}          & 41.82          & 48.89          & 42.85           \\
Gemini-2.5-Flash$^\dagger$                 & 70.18          & 69.30          & 69.87          & 54.66          & 51.89          & 49.80          & 53.21          & 48.22           \\ 
\hline

Ovis2-2B$^\ast$               & 76.97          & 42.48          & 54.82          & 43.98          & 40.83          & 54.51          & 46.69          & 33.16           \\
mPLUG-Owl3-2B                    & 74.86          & 83.18          & 78.88          & 67.59          & 18.35          & 45.99          & 26.37          & 11.61           \\
Qwen2-VL-2B                      & 74.97          & \underline{92.99}  & 83.07          & 74.18          & 23.20          & \underline{76.03}  & 36.65          & 26.55           \\
Qwen2.5-VL-3B                    & 80.35          & 91.41          & 85.57          & 77.98          & 33.75          & 75.82          & 47.60          & 39.56           \\
LLaVA-1.5-7B-hf$^\ast$        & 68.15          & 92.06          & 78.43          & 66.95          & 20.91          & \textbf{81.27} & 33.61          & 22.99           \\
MiMo-7B-RL                       & 80.90          & \textbf{93.46} & 86.71          & 79.78          & 49.46          & 48.22          & 49.05          & 41.84           \\
Qwen2-VL-7B                      & 83.15          & 90.72          & 86.81          & 79.92          & 42.80          & 69.60          & 53.54          & 45.75           \\
Qwen2.5-VL-7B                    & 81.52          & 91.42          & 86.23          & 79.00          & 39.99          & 71.30          & 51.71          & 43.65           \\
InternVL3-8B$^\ast$           & 80.67          & 92.98          & 86.43          & 79.35          & 34.08          & 75.52          & 47.07          & 36.65           \\
DeepSeek-VL2-Small-16B               & 77.52          & 91.13          & 83.82          & 75.23          & 47.43          & 19.48          & 27.91          & 24.51           \\
\rowcolor[rgb]{0.951,0.951,0.951}\textbf{F$^2$RVLM-Qwen2-VL-2B}   & 79.64          & 89.55          & 84.35          & 76.07          & 26.97          & 72.00          & 40.46          & 30.90           \\
\rowcolor[rgb]{0.951,0.951,0.951}\textbf{F$^2$RVLM-Qwen2.5-VL-3B} & 82.74          & 91.65          & 87.00          & 80.19          & 45.24          & 70.86          & 55.60          & 48.09           \\
\rowcolor[rgb]{0.951,0.951,0.951}\textbf{F$^2$RVLM-Qwen2-VL-7B}   & \textbf{84.02} & 90.67          & \textbf{87.25} & \textbf{80.60} & \textbf{57.21} & 67.46          & \textbf{62.07} & \textbf{55.46}  \\
\rowcolor[rgb]{0.951,0.951,0.951}\textbf{F$^2$RVLM-Qwen2.5-VL-7B} & \underline{83.24}  & 91.60          & \underline{87.24}  & \underline{80.55}  & 54.83  & 65.16          & \underline{59.60}  & \underline{52.39}   \\
\bottomrule
\end{tabular}
}
\vspace{-3pt}  
\caption{Comparison with popular VLMs on the MLDR validation set and WeChat test set. ``$\dagger$'' indicates zero-shot inference without MLDR fine-tuning. ``$\ast$'' indicates models limited by context length, evaluated via sliding-window inference.}
\label{tab:2}
\vspace{-15pt}  
\end{table*}

\noindent \textbf{Fragment Order Consistency} $(R_{\text{Fragment}})$.
To encourage semantically coherent, contextually aligned fragment retrieval, we propose a reward based on cross-modal fragment order consistency. It evaluates whether the retrieved utterances and images maintain the natural progression of information, which is especially important in long conversations with intertwined modalities. 
Given the predicted sets of utterance and image IDs, we first locate the corresponding textual and visual elements in the original dialogue and encode them into a unified embedding space with CLIP. These embeddings are then interleaved and temporally ordered according to their original positions within the dialogue, forming a cross-modal sequence $\left \{ e_1,e_2,...,e_{N} \right \} $. The reward is defined as the average pairwise cosine similarity between adjacent embeddings in this sequence:
\begin{equation}
R_{\text{Fragment}} = \frac{1}{N-1} \sum_{i=1}^{N-1} \cos(\mathbf{e}_i, \mathbf{e}_{i+1})
\end{equation}
If the sequence length $N$ is less than 2, a fallback reward (i.e., 0.5) is returned.




\subsection{Difficulty-aware Curriculum Sampling}
Fragment structures in long-form dialogues differ in the distribution of their internal elements: some comprise utterances or images that are tightly clustered within adjacent turns, while others span sparsely across distant positions. This variation significantly affects retrieval difficulty, clustered fragments are generally easier to locate, whereas dispersed ones are harder to handle reliably.


To leverage this inherent difficulty hierarchy, we adopt a difficulty-aware curriculum sampling that dynamically quantifies instance difficulty based on predicted F1 scores and confidence. Training begins with high-confidence, high-F1 samples (i.e., easy cases) and gradually incorporates harder ones with lower scores. This progressive schedule enables the model to first master reasoning in dense contexts, then adapt to long-range, complex scenarios, mirroring the incremental nature of human learning.
Specifically, for each training sample $x_i$, we compute two indicators utilizing a cold-start model: (1) \textbf{Retrieval F1 Score} $f_i$: measures overlap between predicted and ground-truth utterance/image ID sets; and (2) \textbf{Prediction Entropy} $h_i$: quantifies uncertainty as the average entropy of predicted token distributions. Each instance is then assigned a difficulty level based on the following criteria:
\begin{equation}
d_i = 
\begin{cases}
\text{Easy}, & f_i \geq Q_{0.75}^{(f)} \ \text{and} \ h_i \leq Q_{0.25}^{(h)} \\
\text{Confusing}, & f_i \leq Q_{0.25}^{(f)} \ \text{and} \ h_i \geq Q_{0.75}^{(h)} \\
\text{Hard}, & f_i \leq Q_{0.25}^{(f)} \ \text{and} \ h_i \leq Q_{0.25}^{(h)} \\
\text{Medium}, & \text{otherwise}
\end{cases}
\end{equation}
Here, $Q_p^{(f)}$ and $Q_p^{(h)}$ denote the $p$-th percentiles of F1 and entropy distributions. “Easy” samples are confident and correct, “Confusing” are uncertain and incorrect, while “Hard” are incorrect yet overconfident. 




\section{Experiments}

\subsection{Experimental Settings}
\noindent \textbf{Models \& Details.} 
We implement F$^2$RVLM based on the ms-swift~\cite{zhao2024swiftascalablelightweightinfrastructure} framework, using Qwen-VL-series as the backbone with 2B, 3B, and 7B parameters. Parameter-efficient fine-tuning is conducted on the MLDR dataset via LoRA~\cite{hu2022lora}. For evaluation, we compare F$^2$RVLM against a comprehensive set of VLMs, including proprietary models such as GPT-4o~\cite{jaech2024openai}, Gemini-2.5~\cite{comanici2025gemini}, Doubao-Seed-1.6~\cite{guo2025seed1}, and Claude-Sonnet-4; open-source generation-based models including LLaVa~\cite{liu2023llava}, Qwen-VL-series~\cite{wang2024qwen2}, DeepSeek-VL2-series~\cite{wu2024deepseek}, InternVL3-series~\cite{chen2024internvl}, Ovis2~\cite{lu2024ovis}, Owl3~\cite{ye2024mplugowl3longimagesequenceunderstanding}, and Mimo-VL~\cite{coreteam2025mimovltechnicalreport}; as well as embedding-based models such as CLIP~\cite{radford2021learning}, BLIP-2~\cite{li2023blip}, E5-V~\cite{jiang2024e5}, and GME~\cite{zhang2024gme}. Open-source models are fine-tuned on MLDR using SFT, while proprietary and embedding-based models are evaluated in inference-only mode. Full implementation details are provided in the Appendix.\textcolor{red}{B}.


\noindent \textbf{Metrics.}
We evaluate fragment-level retrieval performance utilizing four metrics: Precision, Recall, F1 Score, and Matthews Correlation Coefficient (MCC), computed separately for utterance IDs and image IDs. To obtain unified scores, we average the F1 and MCC values across both modalities and calculate the harmonic mean of Precision and Recall to reflect joint retrieval performance.


\begin{table}[t]
\centering
\resizebox{0.9\linewidth}{!}{
\begin{tabular}{rccc} 
\toprule
Model                                                & \#Params                                                                 & Fragment Consis. & Query Sim.      \\ 
\hline
Doubao-Seed-1.6$^\dagger$                                      & \multirow{4}{*}{\begin{tabular}[c]{@{}c@{}}Closed\\-Source\end{tabular}} & 27.89            & 31.06           \\
Gemini-2.5-Flash$^\dagger$                                     &                                                                          & 37.67            & 41.44           \\
Claude-Sonnet-4$^\dagger$                                      &                                                                          & 42.29            & 46.64           \\
GPT-4o$^\dagger$                                               &                                                                          & 43.17            & 48.21           \\
Qwen2.5-VL$^\dagger$                                           & 72B                                                                      & 45.98            & 48.49           \\ 
\hline
DeepSeek-VL2-Small                                   & 16B                                                                      & 12.52            & 13.67           \\
MiMo-VL-RL                                           & 7B                                                                       & 34.69            & 36.69           \\
Qwen2-VL                                             & 7B                                                                       & \underline{55.28}    & \underline{59.49}   \\
Qwen2.5-VL                                           & 7B                                                                       & 54.81            & 58.39           \\
\rowcolor[rgb]{0.951,0.951,0.951} \textbf{F$^2$RVLM} & 3B                                                                       & \textbf{60.53}   & \textbf{61.18}  \\
\bottomrule
\end{tabular}}
\vspace{-5pt}  
\caption{Comparison with popular VLMs on the real-domain WeChat test set, evaluated by Fragment Order Consistency and Query-Fragment Similarity. ``$\dagger$'' indicates zero-shot inference without MLDR fine-tuning.} \label{tab:3}
\vspace{-5pt}  
\end{table}

\begin{figure}[t]
\centering
\includegraphics[width=0.95\linewidth]{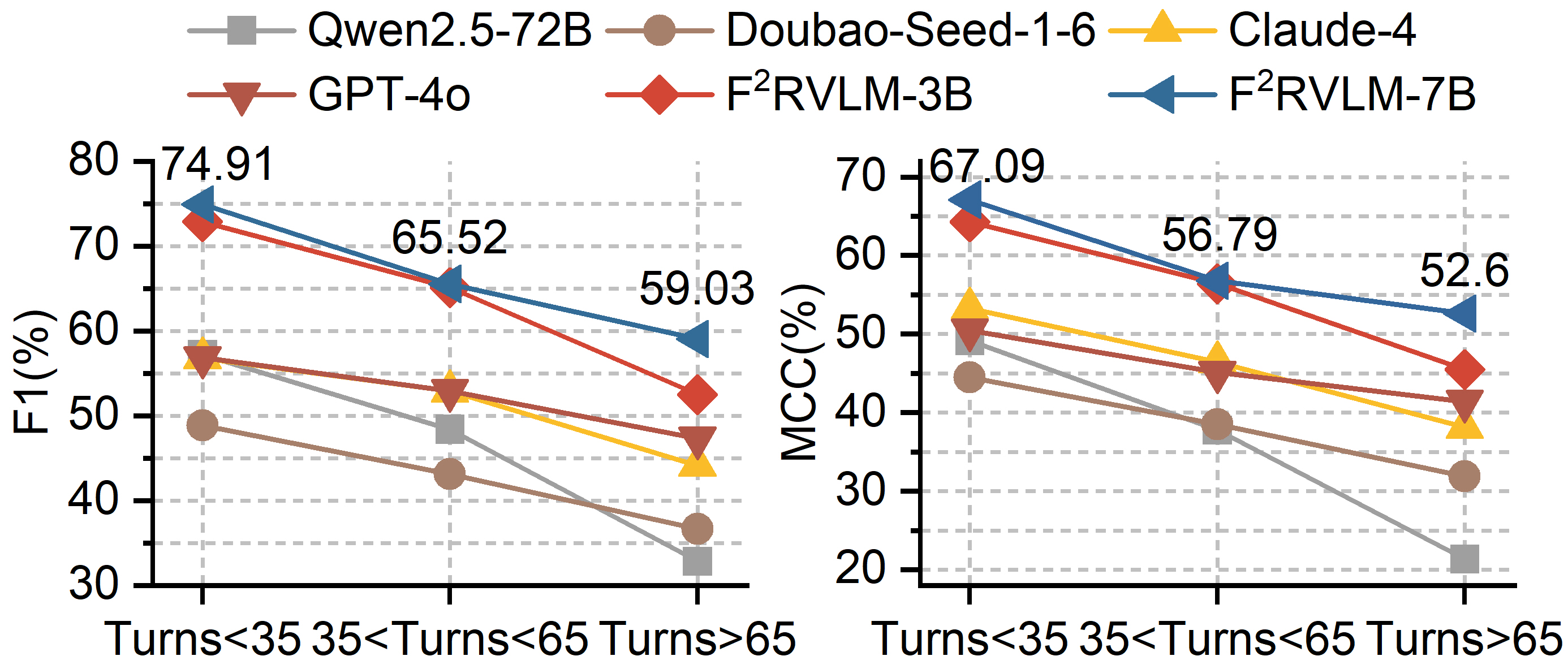}
\vspace{-5pt}  
\caption{Performance comparison across dialogue turn groups on the real-domain WeChat test set.} 
\vspace{-18pt}  
\label{fig:4}
\end{figure}

\subsection{Comparison with Popular VLMs}
\textbf{Overall Retrieval Performance.}
Table~\ref{tab:2} summarizes fragment-level retrieval results on the in-domain MLDR validation set and the real-domain WeChat test set. Key observations include: 
(1) F$^2$RVLM achieves SOTA performance in both domains. The 7B variant tops MLDR with 87.25\% F1, and leads on WeChat with 62.07\% F1. Its 3B version also outperforms larger competitors such as MiMo-7B-RL, and GPT-4o.
(2) MLDR Fine-tuning significantly boosts cross-domain retrieval.
Qwen2.5-VL-7B improves from 12.58\% (zero-shot) to 51.71\% F1 on WeChat after MLDR tuning, demonstrating MLDR's value as a supervised resource for fragment retrieval. Moreover, F$^2$RVLM offers less performance degradation from MLDR to WeChat compared to other models, indicating stronger generalization in real-world long-form dialogues. (3) Detailed metric results, qualitative analysis, and further discussion on generalization can be found in Appendix.\textcolor{red}{C}.

\noindent \textbf{Discussion about Recall Metric.} While models like LLaVA achieve higher recall, they often sacrifice precision by retrieving many irrelevant fragments. In contrast, our model incorporates a penalty term in the $R_{\text{F1}}$ to suppress over-retrieval, encouraging the selection of fewer but more semantically consistent fragments, better aligned with human preferences and fragment-level retrieval objectives.

\begin{figure}[t]
\centering
\includegraphics[width=0.8\linewidth]{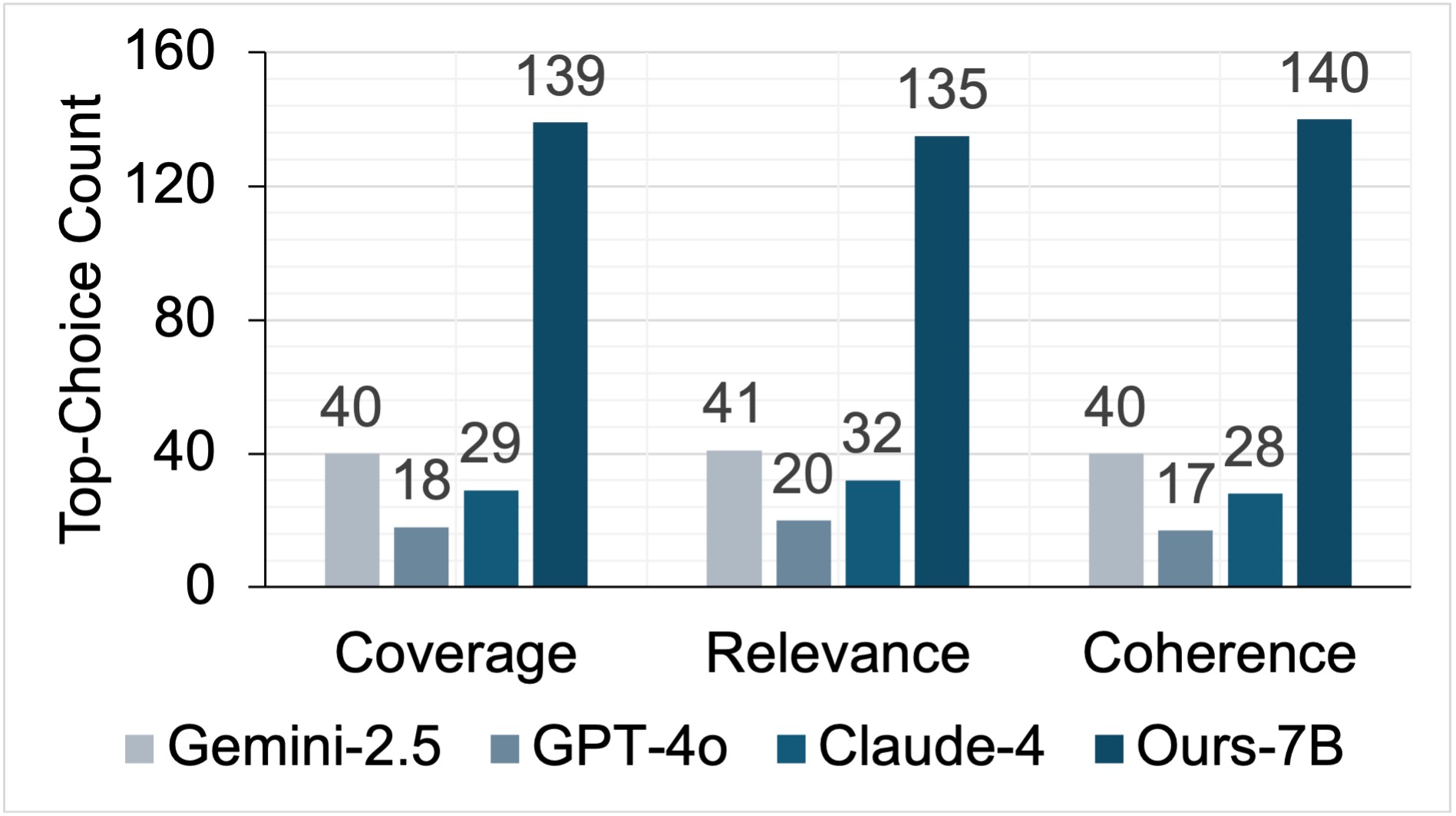}
\vspace{-5pt}  
\caption{Human subjective evaluation results on the WeChat test set. We randomly sample 200 dialogues and ask expert annotators to select the top-performing model across three criteria: Coverage, Relevance, and Coherence.}
\vspace{-5pt}  
\label{fig:5}
\end{figure}

\begin{table}[t]
\centering
\resizebox{0.95\linewidth}{!}{
\begin{tabular}{ccc|cccc} 
\toprule
\multirow{2}{*}{$R_\text{F1}$}         & \multirow{2}{*}{$R_\text{Fragment}$}   & \multirow{2}{*}{Curriculum}            & \multicolumn{2}{c}{In-domain} & \multicolumn{2}{c}{Real-domain}  \\
&                                        &                                        & F1(\%)         & MCC(\%)             & F1(\%)         & MCC(\%)                 \\ 
\hline
\textcolor[rgb]{0.702,0.702,0.702}{\ding{55}} & \textcolor[rgb]{0.702,0.702,0.702}{\ding{55}} & \ding{51}                                     & 86.02          & 78.69               & 53.39          & 46.07                   \\
\ding{51}                                     & \textcolor[rgb]{0.702,0.702,0.702}{\ding{55}} & \ding{51}                                     & \underline{86.39}  & \underline{79.24}       & \underline{54.68}  & \underline{47.41}           \\
\ding{51}                                     & \ding{51}                                     & \textcolor[rgb]{0.702,0.702,0.702}{\ding{55}} & 84.69          & 77.53               & 50.65          & 43.71                   \\
\ding{51}                                     & \ding{51}                                     & \ding{51}                                     & \textbf{87.00} & \textbf{80.19}      & \textbf{55.60} & \textbf{48.09}          \\
\bottomrule
\end{tabular}}
\vspace{-3pt}  
\caption{Ablation study on reward components and curriculum sampling on MLDR validation and WeChat test sets.} \label{tab:5}
\vspace{-15pt}  
\end{table}


\noindent \textbf{Fragment-level Consistency and Alignment.}
To assess VLMs’ ability to capture dialogue structure and semantic alignment, we introduce two metrics on the WeChat test set: Fragment Order Consistency (average cosine similarity between adjacent retrieved elements) and Query-Fragment Similarity (average similarity between the query and each retrieved element), as reported in Table~\ref{tab:3}. F$^2$RVLM achieves the highest scores on both, outperforming larger open- and closed-source models. This demonstrates its superior ability to retrieve coherent, semantically aligned fragments, enabled by the order-consistency reward in GRPO training.


\noindent \textbf{Discussion about Dialogue Turns.}
Fig.\ref{fig:4} reports model performance on the WeChat test set across short ($<$35 turns), medium (35–65 turns), and long ($>$65 turns) dialogues. F$^2$RVLM-7B (Qwen2) consistently achieves the highest F1 and MCC scores as dialogue turn increases, with minimal performance degradation compared to other models, which demonstrates its robustness in long-context reasoning.


\noindent \textbf{Subjective Results.}
We conduct a human evaluation on 200 dialogues from the WeChat test set, comparing our 7B model against Gemini-2.5, GPT-4o, and Claude-4 across three criteria: coverage, relevance, and fragment coherence (detailed definitions are provided in Appendix.\textcolor{red}{C}). As depicted in Fig.\ref{fig:5}, our model is consistently preferred by annotators, receiving the highest top-choice counts in all aspects, demonstrating superior semantic completeness, alignment, and fluency in real-world dialogue retrieval.

\subsection{Ablation Results}
This section presents ablation studies to evaluate the effectiveness of the proposed reward functions and personalized curriculum learning, focusing on joint F1 and MCC metrics (Table~\ref{tab:5}). (1) Replacing standard accuracy with $R_{\text{F1}}$ significantly improves performance, while incorporating fragment order consistency $R_{\text{Fragment}}$ further enhances cross-domain F1 and MCC, highlighting the role of intra-fragment semantic consistency. (2) Integrating difficulty-aware curriculum sampling into GRPO yields consistent gains: in-domain F1 improves from 84.69\% to 87.00\%, and real-domain F1 from 50.65\% to 55.60\%, validating the effectiveness of progressive learning from easier to harder samples. (3) We also explore different training ratios of SFT and GRPO-based RFT, with detailed results provided in Appendix.\textcolor{red}{C}.

\section{Conclusion}
This study defines Fine-grained Fragment Retrieval (FFR) to locate semantically coherent utterance-image fragments from long-form multimodal dialogues. To support this task, we construct MLDR, the longest-turn multimodal dialogue dataset to date, along with a real-world WeChat test set. Based on these resources, we propose F$^2$RVLM, a generation-based retrieval model trained with GRPO-based reinforcement learning to encourage semantically coherent fragment prediction. F$^2$RVLM surpasses popular VLMs, achieving superior accuracy in both in-domain and real-world evaluations.



\bigskip

\bibliography{aaai2026}
\clearpage
\appendix

\section{Appendix Overview}

The Supplementary Materials offer additional insights and experimental details that support the main paper. The content is organized as follows:

\begin{itemize}
\item \textbf{Appendix A: Dataset Construction and Analysis}
\begin{itemize}
\item \textbf{A1. MLDR Construction:} Describes the full pipeline for building the MLDR dataset.
\item \textbf{A2. WeChat-Based Dataset Construction:} Details the data source, annotation protocol, and licensing for the WeChat test set.
\item \textbf{A3. Dataset Statistics and Analysis:} Presents dialogue turn distribution, topic bias analysis, and representative qualitative examples.
\end{itemize}

\item \textbf{Appendix B: Implementation Details}
\begin{itemize}
\item \textbf{B1. Training and Hyperparameter Settings:} Summarizes training configurations and optimization details.
\item \textbf{B2. Model Comparison Settings:} Specifies the fine-tuning and inference setups for all models, including both open-source and proprietary VLMs.
\item \textbf{B3. Sliding Window Inference for Long Contexts:} Explains inference strategies for models that cannot process long dialogues directly.
\end{itemize}

\item \textbf{Appendix C: Experimental Results and Analysis}
\begin{itemize}
\item \textbf{C1. Detailed Comparison of Quantitative Results:} Provides full metric breakdowns on MLDR and WeChat datasets.
\item \textbf{C2. Comparison of Qualitative Results:} Compares typical retrieval examples across models.
\item \textbf{C3. Cross-Lingual Generalization:} Evaluates performance on an English-translated version of the WeChat test set.
\item \textbf{C4. Human Subjective Evaluation Criteria:} Defines coverage, relevance, and coherence metrics used in human evaluation.
\item \textbf{C5. Ablation Studies on SFT and GRPO Ratio:} Analyzes the impact of different supervised and reinforcement fine-tuning ratios.
\end{itemize}
\end{itemize}

\section{Appendix A: Dataset Construction and Analysis}

\subsection{A1. MLDR Construction}
This section provides detailed elaboration on the key steps involved in the construction of the MLDR dataset.

\noindent \textbf{Data Cleaning.}
To ensure high-quality long-form construction, we conduct a rigorous cleaning process based on four key filtering standards: (1) Image-Text Consistency: We encode visual and textual content utilizing CLIP and discard samples with low cross-modal similarity. (2) Topic Coherence: Sentence embeddings are obtained utilizing BERT, and K-means clustering is applied to filter out dialogues that deviate from a central topic. (3) Dialogue Turn Structure: Each dialogue must contain at least three turns in a ``User1–User2–User1" pattern to ensure basic conversational flow. (4) Image Quality: We filter out images with resolutions below 500 pixels or with extreme aspect ratios (greater than 7.5), ensuring visual quality.

\begin{table*}[t]
\centering
\resizebox{1.0\linewidth}{!}{
\begin{tabular}{c|cccc|ccccc} 
\toprule
\textbf{Datasets}                                      & \textbf{Modalites} & \textbf{Dialogue Type} & \textbf{Dialogue Source} & \textbf{Language} & \textbf{Dialogues} & \textbf{Images} & \textbf{Turns} & \textbf{Turn/Dialog} & \textbf{Topic/Dialog}  \\ 
\hline
ImageChat~\cite{shuster2020image}                                              & v, t               & image-grounded         & crowdsourcing            & English           & 201,779            & 201,779         & 400,853        & 1.98~                & 1.00~                  \\
OpenViDial~\cite{meng2020openvidial}                                             & v, t               & image-grounded         & moviesTVs                & English           & 1,100,000          & 1,100,000       & 1,100,000      & 1.00~                & 1.00~                  \\
PhotoChat~\cite{zang2021photochat}                                              & v, t               & image-sharing          & crowdsourcing            & English           & 11,820             & 10,479          & 150,138        & 12.74~               & 1.00~                  \\
MMDD~\cite{lee2021constructing}                                                   & v, t               & image-sharing          & text datasets            & English           & 17,679             & 13,288          & 187,421        & 11.56~               & 1.00~                  \\
MMDialog~\cite{feng2023mmdialog}                                               & v, t               & image-sharing          & social media             & English           & 1,079,117          & 1,556,868       & 4,920,000      & 4.56~                & 1.00~                  \\
DialogCC~\cite{lee2024dialogcc}                                               & v, t               & image-sharing          & text datasets            & English           & 83,209             & 129,802         & 676,181        & 8.20~                & 1.00~                  \\
\hline
MMChat~\cite{zheng2022mmchat}                                                 & v, t               & image-grounded         & social media             & Chinese           & 120,840            & 204,320         & 314,130        & 2.59~                & 1.00~                  \\

M3ED~\cite{zhao2022m3ed}                                                   & a, v, t            & video-grounded         & TVs                      & Chinese           & 990                & -               & 9082           & 9.17~                & 1.00~                  \\
CPED~\cite{chen2022cped}                                                   & a, v, t            & video-grounded         & TVs                      & Chinese           & 12,000             & -               & 133,000        & 11.08~               & 1.00~                  \\
CMMA~\cite{zhang2023cmma}                                                   & a, v, t            & video-grounded         & TVs                      & Chinese           & 3,000              & -               & 21,795         & 7.27~                & 1.00~                  \\
TikTalk~\cite{lin2023tiktalk}                                                & a, v, t            & video-grounded         & social media             & Chinese           & 367,670            & 38,703 videos   & 826,752        & 2.25~                & 1.00~                  \\ 
\hline
\rowcolor[rgb]{0.951,0.951,0.951} \textbf{Our Dataset} & v, t               & image-sharing          & text datasets            & Chinese           & 37,030             & 194,543         & 942,414        & \textbf{25.45}       & \textbf{3.00}          \\
\bottomrule
\end{tabular}}
\caption{Summary of main multimodal dialogue datasets. Datasets are categorized by modality (audio~(a), visual~(v), and textual~(t)), dialogue type (image/video-grounded vs. image-sharing), source, and language. Our MLDR dataset features significantly longer dialogues and richer topic transitions, better reflecting real-world human communication scenarios.} \label{tab:A1}
\end{table*}

\begin{figure*}[htbp]
\centering
\includegraphics[width=1\linewidth]{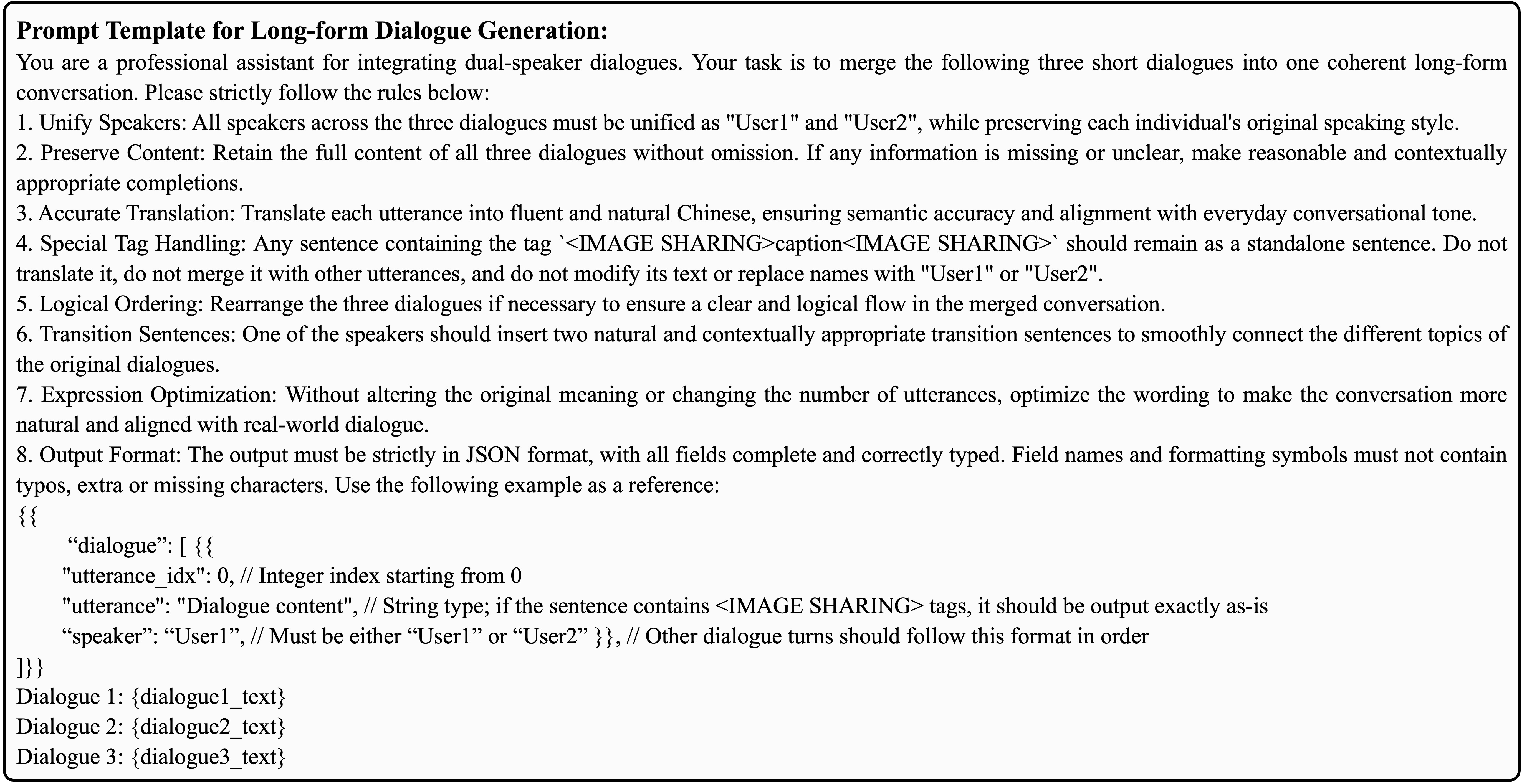}
\caption{Prompt Template for Long-form Dialogue Generation.}
\label{fig:A1}
\end{figure*}

\begin{figure*}[!ht]
\centering
\includegraphics[width=0.9\linewidth]{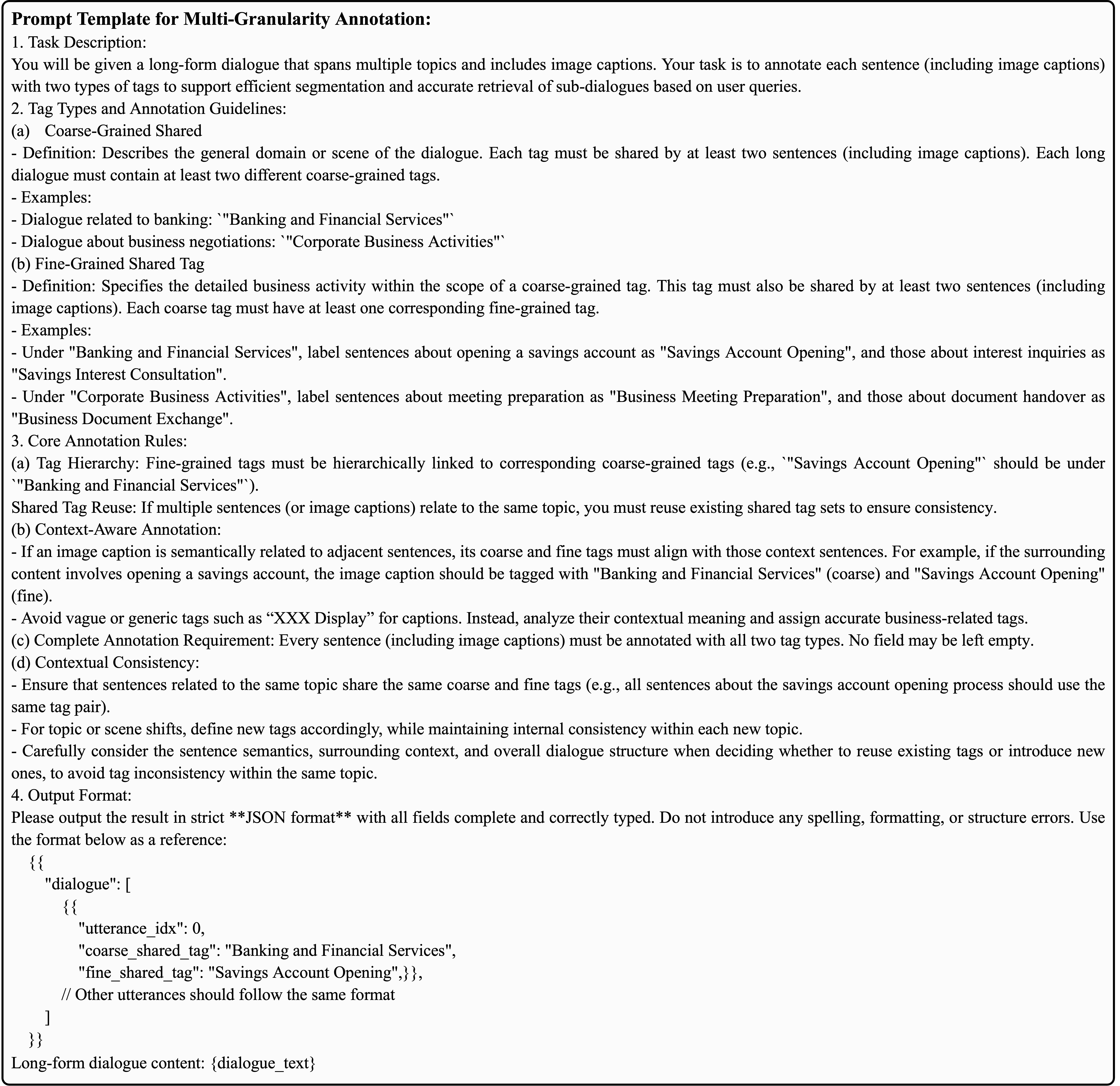}
\caption{Prompt Template for Multi-Granularity Annotation.}
\label{fig:A2}
\end{figure*}

\noindent \textbf{Dialogue Triplet Semantic Matching.}
To generate coherent long-form dialogues from multiple short dialogues, we propose a three-step triplet-based semantic matching strategy. This process ensures both topic continuity and multimodal alignment, and includes the following steps:
\begin{itemize}
\item Trial Screening of Text Semantic: For each cleaned short dialogue \( D_A \), we first perform textual semantic screening. We encode all textual content (including image captions) using a pretrained BERT model and compute the mean sentence embedding for each dialogue. Cosine similarity is then calculated between \( D_A \)’s embedding and the embeddings of all other candidate dialogues. Based on this initial similarity score, we select the Top-\(K\) (where \(K = 50\)) most semantically relevant dialogues.
\item Refined Screening of Multimodal Semantic: Next, we perform multimodal semantic refinement for the Top-\(K\) dialogues selected in the first step. Using CLIP, we separately encode both textual and visual content, calculating the cosine similarity for each modality. These similarities are then combined with weighted scores (i.e., 0.7 for text and 0.3 for image) to obtain the final multimodal similarity score. From this, we select the Top-2 dialogues \( \{D_B^1, D_B^2\} \), based on this refined multimodal score.
\item Triplet Construction: For each selected dialogue \( D_B^i \) (\(i = 1, 2\)), we repeat the above two-stage matching process to retrieve the Top-2 most semantically and multimodally relevant dialogues, denoted as \( \{D_C^{i,1}, D_C^{i,2}\} \). This results in 4 distinct triplets: 
\begin{align} \label{equation:1}
D_A \rightarrow D_B^i \rightarrow D_C^{i,j}, \quad i = 1,2; \quad j = 1,2
\end{align}

These triplets provide a structural foundation for generating extended dialogues, ensuring both semantic and multimodal coherence. Notably, each dialogue \( D_A \), \( D_B \), and \( D_C \) in a triplet must be mutually disjoint in terms of both dialogue ID and image content, ensuring that each triplet represents unique and non-duplicate information.
\end{itemize}

\noindent \textbf{Long-form Dialogue Generation.} 
To convert semantically aligned triplets into coherent long-form dialogues, we employ the Qwen3 256B~\cite{yang2025qwen3} large language model under a structured prompt template (Fig.\ref{fig:A1}). The generation process adheres to the following principles: (1) Preserve the original multimodal content and semantics while enhancing fluency and readability. (2) Insert natural transition utterances to ensure coherent topic progression and stylistic consistency across dialogue segments. (3) Translating all content into fluent and contextually appropriate Chinese, aligned with real-world conversational norms. This process enables the synthesis of high-quality, semantically coherent, and structurally complete Chinese multimodal long-form dialogues, which form the foundation for subsequent retrieval and understanding tasks.

\noindent \textbf{Multi-Granularity Annotation.} 
To enable fine-grained semantic retrieval over long-form multimodal dialogues, we design a two-level shared tagging scheme and implement it automatically using a Qwen3 model under structured prompts (see Fig.\ref{fig:A2}). Each sentence and image caption is annotated with (1) a coarse-grained shared tag indicating the high-level domain or scenario (e.g., “banking services”, “travel planning”), and (2) a fine-grained shared tag specifying the detailed activity within the domain (e.g., “open savings account”). Both types of tags are required to be shared by at least two utterances or image captions to ensure topic consistency. To support high-quality tagging, Qwen3 is guided by explicit rules including tag reuse, topic-aligned grouping, and context-aware alignment across dialogue turns. This automated annotation process enables efficient semantic disentanglement and facilitates precise retrieval in complex multimodal conversations.

\subsection{A2. WeChat-Based Dataset Construction}
\textbf{Data Source and Licensing Statement.}
The real-domain WeChat test set is constructed from naturally occurring image-text conversations voluntarily contributed by 12 participants. These dialogues are not extracted from any internal WeChat databases, but are shared with explicit informed consent from the contributors for academic research purposes. Prior to inclusion, all conversations were thoroughly anonymized and sanitized, with personal identifiers, sensitive content, and inappropriate language removed to ensure privacy protection and ethical compliance. To support responsible use, the dataset will be released under a research-only license. It is strictly prohibited to use the data for commercial applications, product development, or any non-academic purposes. Access to the dataset requires agreement to the accompanying license terms.

\noindent \textbf{Annotation Protocol and Quality Control.}
To ensure high-quality supervision for real-world fragment retrieval, we recruited 10 trained annotators to label the WeChat test set. Prior to annotation, all annotators received standardized training covering task definitions, multi-granularity labeling rules, and edge case handling. Each dialogue was independently annotated by all annotators to ensure consistency and comprehensive coverage. Following the initial round, we conducted rigorous quality checks to identify inconsistencies and labeling errors. Annotators were required to revise their annotations based on feedback. This iterative process continued until all annotations passed final validation, resulting in a reliable and well-curated benchmark for evaluating fine-grained multimodal retrieval.

\begin{figure}[t]
\centering
\includegraphics[width=1.0\linewidth]{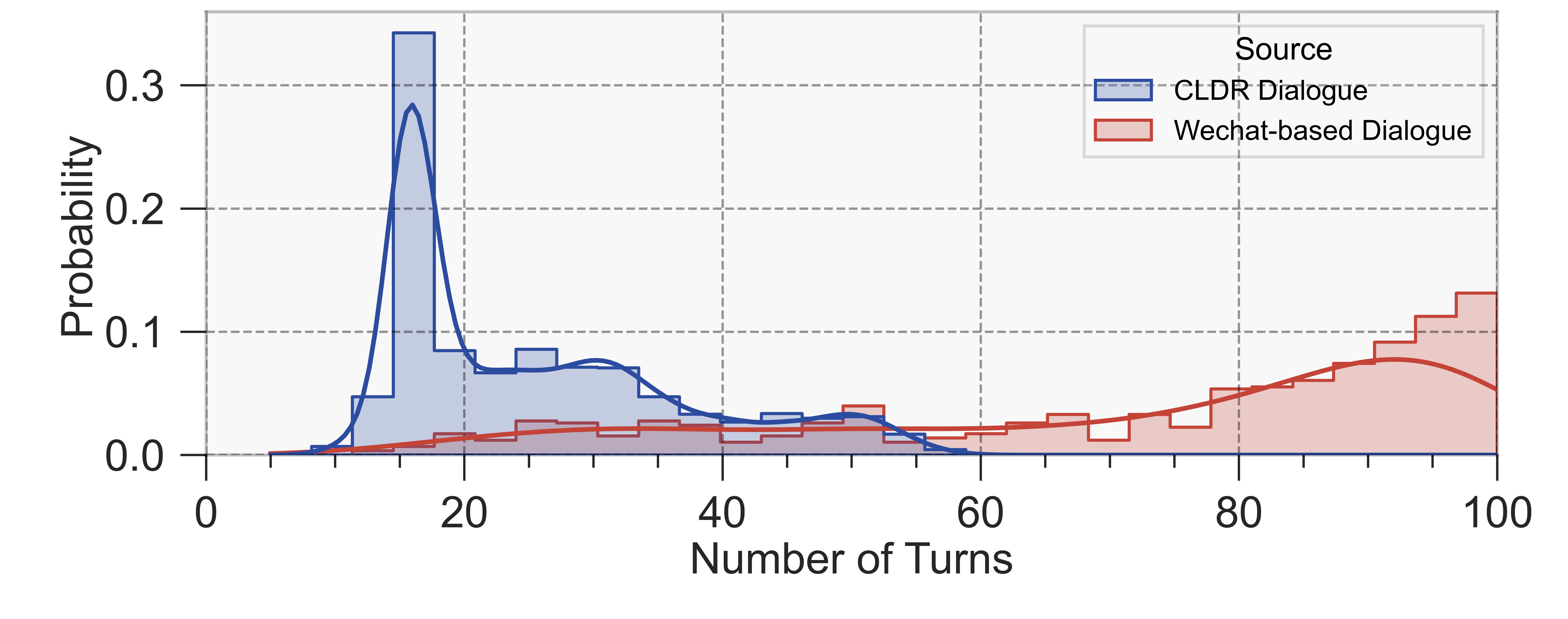}
\vspace{-15pt}  
\caption{Comparison between the MLDR and WeChat-based datasets in terms of dialogue turn distributions.}
\vspace{-15pt}  
\label{fig:A3}
\end{figure}

\subsection{A3. Dataset Statistics and Analysis}
\textbf{Dataset Cases.}
To illustrate the characteristics of different evaluation scenarios, we present representative cases from both the MLDR and the WeChat test set (as reported in Fig.\ref{fig:A4}-\ref{fig:A6}). The MLDR set contains dialogues with fewer turns and a cleaner structure, while the WeChat test set features long, real-world conversations with many turns and mixed topics, posing greater challenges for retrieval models in maintaining relevance and coherence.

\noindent \textbf{Topic Distribution Bias in the WeChat Test Set.}
As illustrated in Fig.3 in the main body, the WeChat test set exhibits a pronounced skew toward Work \& Tech topics, which account for nearly 50\% of all annotated dialogues. This bias can be primarily attributed to two factors. (1) The dataset is constructed from real-world conversations voluntarily contributed by 12 participants, most of whom are doctoral students or corporate employees. Their daily communication naturally revolves around professional and technical discussions, resulting in a topic distribution that overrepresents work-related content. (2) In compliance with ethical and privacy considerations, we intentionally filtered out sensitive or highly private conversations (e.g., involving family, personal health, or intimate relationships). This further reduces the proportion of daily-life and emotion-related content.

We acknowledge that this topic bias represents a limitation of the current dataset. While it provides valuable insights into real-world retrieval performance under noisy and task-oriented scenarios, it may not fully capture the diversity of open-domain multimodal interactions. In future iterations, we plan to expand the contributor pool to include participants from more diverse demographics and occupations, and to adopt privacy-preserving mechanisms (e.g., local anonymization) that allow for the safe inclusion of a wider range of topics. This will support the construction of a more balanced and comprehensive real-domain benchmark.

\noindent \textbf{Dialogue Turn Distribution.}
As illustrated in Fig.\ref{fig:3}, the MLDR and WeChat test sets exhibit markedly different distributions in the number of dialogue turns. The MLDR dialogues are densely concentrated around 15–25 turns, with a sharp peak at around 20, reflecting the synthetic construction of the dataset via triplet concatenation under controlled prompting. This design ensures topic continuity and moderate length, which is beneficial for training and controlled evaluation. In contrast, the WeChat-based dialogues demonstrate a broad and skewed distribution, ranging from short to extremely long conversations, with a significant proportion exceeding 60 turns and some approaching 100. This reflects the organic nature of real-world dialogues, which often involve prolonged multi-topic interactions with informal exchanges and irregular structure. The longer and noisier dialogue structure poses greater challenges for retrieval, especially in maintaining semantic consistency and aligning queries to sparse but relevant fragments.

\begin{table*}[t]
\centering
\resizebox{1.0\linewidth}{!}{
\begin{tabular}{r|cccc|cccc|cccc} 
\toprule
\multirow{2}{*}{Model}                                          & \multicolumn{4}{c|}{Utterance Retrieval}                          & \multicolumn{4}{c|}{Image Retrieval}                              & \multicolumn{4}{c}{Joint Retrieval}                                \\ 
\cline{2-13}
& Precision      & Recall         & F1             & MCC            & Precision      & Recall         & F1             & MCC            & Precision      & Recall         & F1             & MCC             \\ 
\hline

CLIP-Embedding$^\dagger$                    & 45.21          & 20.42          & 28.13          & 12.18          & 52.86          & 62.63          & 57.33          & 30.11          & 48.74          & 30.80          & 42.73          & 21.14           \\
BLIP2-Embedding$^\dagger$                   & 30.54          & 78.97          & 44.05          & 0.00           & 33.33          & 1.51           & 2.89           & 0.00           & 31.88          & 2.96           & 23.47          & 0.00            \\
E5-V-Embedding$^\dagger$                    & 54.39          & 40.16          & 46.21          & 27.31          & 53.32          & 62.25          & 57.44          & 30.54          & 53.85          & 48.83          & 51.82          & 28.92           \\
GME-Embedding$^\dagger$                     & 56.96          & 33.09          & 41.86          & 26.09          & 68.66          & 18.52          & 29.18          & 22.25          & 62.27          & 23.75          & 35.52          & 24.17           \\
Qwen2.5-VL-7B$^\dagger$                     & 23.29          & 3.08           & 5.45           & 0.00           & 19.46          & 6.95           & 10.24          & 0.00           & 21.20          & 4.27           & 7.84           & 0.00            \\
MiMo-7B-RL$^\dagger$                        & 70.18          & 46.23          & 55.74          & 42.74          & 65.35          & 68.44          & 66.86          & 48.30          & 67.68          & 55.19          & 61.30          & 45.52           \\
Qwen2.5-VL-72B$^\dagger$                    & 62.42          & 60.67          & 61.54          & 44.27          & 59.85          & 75.15          & 66.63          & 44.95          & 61.11          & 67.14          & 64.09          & 44.61           \\
Doubao-Seed-1.6$^\dagger$                   & 79.69          & 33.69          & 47.36          & 40.79          & 68.78          & 56.42          & 61.99          & 44.36          & 73.83          & 42.19          & 54.67          & 42.57           \\
Claude-Sonnet-4$^\dagger$                   & 69.88          & 48.51          & 57.26          & 43.83          & 64.74          & 72.37          & 68.34          & 49.31          & 67.21          & 58.09          & 62.80          & 46.57           \\
GPT-4o$^\dagger$                            & 73.90          & 46.51          & 57.09          & 46.11          & 67.28          & 60.24          & 63.56          & 45.70          & 70.43          & 52.49          & 60.32          & 45.91           \\ 
Gemini-2.5-Flash$^\dagger$                  & 68.85          & 64.85          & 66.79          & 52.27          & 71.56          & 74.42          & 72.96          & 57.06          & 70.18          & 69.30          & 69.87          & 54.66           \\
\hline
InternVL3-1B                                                    & 62.49          & 89.16          & 73.48          & 60.64          & 72.45          & 90.63          & 80.53          & 68.65          & 67.10          & 89.89          & 77.00          & 64.64           \\
InternVL3-2B                                                    & 72.68          & 91.31          & 80.94          & 71.84          & 79.04          & 92.41          & 85.20          & 76.38          & 75.73          & 91.86          & 83.07          & 74.11           \\
Ovis2-2B                                                        & 73.85          & 46.27          & 56.90          & 45.33          & 80.37          & 39.26          & 52.75          & 42.63          & 76.97          & 42.48          & 54.82          & 43.98           \\
mPLUG-Owl3-2B                                                   & 71.70          & 81.80          & 76.42          & 64.93          & 78.32          & 84.60          & 81.34          & 70.25          & 74.86          & 83.18          & 78.88          & 67.59           \\
Qwen2-VL-2B                                                     & 71.88          & \underline{92.66}  & 80.96          & 71.97          & 78.35          & 93.32          & 85.18          & 76.38          & 74.97          & 92.99          & 83.07          & 74.18           \\
Qwen2.5-VL-3B                                                   & 77.21          & 91.22          & 83.64          & 75.85          & 83.75          & 91.60          & 87.50          & 80.12          & 80.35          & 91.41          & 85.57          & 77.98           \\
DeepSeek-VL2-Tiny-3B                                            & 58.54          & 83.69          & 68.89          & 52.87          & 68.12          & 84.81          & 75.56          & 60.30          & 62.97          & 84.25          & 72.23          & 56.59           \\
\rowcolor[rgb]{0.951,0.951,0.951} \textbf{F$^2$RVLM-Qwen2-VL-2B}   & 76.84          & 88.86          & 82.42          & 73.97          & 82.64          & 90.25          & 86.28          & 78.16          & 79.64          & 89.55          & 84.35          & 76.07           \\
\rowcolor[rgb]{0.951,0.951,0.951} \textbf{F$^2$RVLM-Qwen2.5-VL-3B} & 80.71          & 90.49          & 85.32          & 78.35          & 84.88          & 92.84          & 88.68          & 82.03          & 82.74          & 91.65          & 87.00          & 80.19           \\ 
\hline
LLaVa-1.5-7B-hf                                                 & 63.99          & 91.93          & 75.46          & 63.72          & 72.88          & 92.19          & 81.41          & 70.18          & 68.15          & 92.06          & 78.43          & 66.95           \\

MiMo-7B-SFT                                                     & 79.01          & 92.55          & 85.25          & 78.28          & 84.38          & \underline{93.97}  & \underline{88.92}  & \underline{82.41}  & 81.61          & \underline{93.25}  & 87.08          & 80.35           \\
MiMo-7B-RL                                                      & 78.17          & \textbf{92.85} & 84.88          & 77.75          & 83.63          & \textbf{94.08} & 88.55          & 81.81          & 80.90          & \textbf{93.46} & 86.71          & 79.78           \\
Qwen2-VL-7B                                                     & 80.60          & 89.60          & 84.86          & 77.66          & 85.86          & 91.87          & 88.76          & 82.18          & 83.15          & 90.72          & 86.81          & 79.92           \\
Qwen2.5-VL-7B                                                   & 78.74          & 90.88          & 84.37          & 76.93          & 84.51          & 91.98          & 88.09          & 81.07          & 81.52          & 91.42          & 86.23          & 79.00           \\
InternVL3-8B                                                    & 77.74          & 92.48          & 84.47          & 77.14          & 83.82          & 93.48          & 88.39          & 81.56          & 80.67          & 92.98          & 86.43          & 79.35           \\
DeepSeek-VL2-Small-16B                                          & 74.93          & 90.56          & 82.01          & 73.40          & 80.29          & 91.71          & 85.62          & 77.05          & 77.52          & 91.13          & 83.82          & 75.23           \\
\rowcolor[rgb]{0.951,0.951,0.951} \textbf{F$^2$RVLM-Qwen2-VL-7B}   & \textbf{82.19} & 88.99          & \underline{85.46}  & \underline{78.58}  & \textbf{85.93} & 92.41          & \textbf{89.05} & \textbf{82.63} & \textbf{84.02} & 90.67          & \textbf{87.25} & \textbf{80.60}  \\
\rowcolor[rgb]{0.951,0.951,0.951} \textbf{F$^2$RVLM-Qwen2.5-VL-7B} & \underline{81.64}  & 90.49          & \textbf{85.84} & \textbf{79.12} & \underline{84.91}  & 92.73          & 88.65          & 81.97          & \underline{83.24}  & 91.60          & \underline{87.24}  & \underline{80.55}   \\
\bottomrule
\end{tabular}
}
\vspace{-3pt}  
\caption{Comparison with popular VLMs on the MLDR validation set. We report Precision, Recall, F1, and MCC for utterance, image, and joint predictions. ``$\dagger$'' indicates zero-shot inference without MLDR fine-tuning.}
\label{tab:A2}
\end{table*}

\begin{table*}[t]
\centering
\resizebox{1.0\linewidth}{!}{
\begin{tabular}{r|cccc|cccc|cccc} 
\toprule
\multirow{2}{*}{Model}                                         & \multicolumn{4}{c|}{Utterance Retrieval}                                         & \multicolumn{4}{c|}{Image Retrieval}                                        & \multicolumn{4}{c}{Joint Retrieval}                                          \\ 
\cline{2-13}
& Precision      & Recall         & F1             & MCC            & Precision      & Recall         & F1             & MCC            & Precision      & Recall         & F1             & MCC             \\ 
\hline
CLIP-Embedding$^\dagger$                    & 16.44          & 35.99          & 22.57          & 9.15           & 27.01          & 81.37          & 40.55          & 26.08          & 20.44          & 49.91          & 31.56          & 17.62           \\
BLIP2-Embedding$^\dagger$                   & 11.00          & 71.41          & 19.07          & 0.00           & 24.68          & 41.81          & 31.04          & 11.97          & 15.22          & 52.74          & 25.05          & 4.94            \\
E5-V-Embedding$^\dagger$                    & 34.73          & 35.82          & 35.27          & 26.80          & 26.92          & 68.89          & 38.72          & 22.00          & 30.33          & 47.13          & 36.99          & 24.40           \\
GME-Embedding$^\dagger$                     & 32.54          & 39.12          & 35.52          & 26.51          & 27.19          & 82.67          & 40.93          & 26.86          & 29.63          & 53.11          & 38.22          & 26.68           \\
Qwen2.5-VL-7B$^\dagger$                                     & 13.95          & 11.38          & 12.53          & 2.77           & 8.07           & 28.98          & 12.62          & 0.00           & 10.22          & 16.34          & 12.58          & 0.00            \\
MiMo-7B-RL$^\dagger$                                     & 41.63          & 22.70          & 29.38          & 24.49           & 41.85           & 26.55          & 32.49          & 22.88           & 41.74          & 24.48          & 30.94          & 23.68            \\
Qwen2.5-VL-72B$^\dagger$                                    & 28.25          & 27.90          & 28.07          & 18.88          & 38.41          & 54.70          & 45.13          & 31.63          & 32.55          & 36.95          & 36.60          & 25.26           \\ 
Doubao-Seed-1.6$^\dagger$                                    & 61.83          & 21.81          & 32.24          & 32.49          & 49.89          & 41.01          & 45.02          & 35.62          & 55.22          & 28.47          & 38.63          & 34.06           \\
Claude-Sonnet-4$^\dagger$                                    & 60.38          & 29.89          & 39.98          & 37.69          & 44.36          & 64.68          & 52.63          & 43.31          & 51.15          & 40.88          & 46.30          & 40.50           \\
GPT-4o$^\dagger$                                             & 60.43          & 32.19          & 42.01          & 39.18          & 52.37          & 59.65          & 55.77          & 46.52          & \underline{56.11}          & 41.82          & 48.89          & 42.85           \\
Gemini-2.5-Flash$^\dagger$                                   & 48.65          & 36.86          & 41.94          & 35.92          & 55.60          & 76.76          & \underline{64.49}          & \textbf{60.52}          & 51.89          & 49.80          & 53.21          & 48.22           \\

\hline

InternVL3-1B$^\ast$                                                  & 19.97          & \underline{86.85} & 32.48          & 15.27          & 29.69          & \underline{73.92} & 42.37          & 41.39          & 23.88          & \underline{79.86} & 37.42          & 28.33           \\
InternVL3-2B$^\ast$                                                  & 28.34          & 81.31          & 42.02          & 22.41          & 35.70          & 67.18          & 46.62          & 44.27  & 31.59          & 73.57          & 44.32          & 33.34           \\
Ovis2-2B$^\ast$                                                      & 39.72  & 56.87          & 46.77  & 23.54          & 42.00          & 52.34          & 46.60          & 42.79          & 40.83          & 54.51          & 46.69          & 33.16           \\
mPLUG-Owl3-2B                                                        & 20.65          & 45.01          & 28.31          & 16.78          & 16.51          & 47.01          & 24.44          & 6.44           & 18.35          & 45.99          & 26.37          & 11.61           \\
Qwen2-VL-2B                                                   & 18.38          & 86.49  & 30.32          & 23.64          & 31.46          & 67.83          & 42.98          & 29.45          & 23.20          & 76.03  & 36.65          & 26.55           \\
Qwen2.5-VL-3B                                                 & 27.77          & 79.03          & 41.11          & 35.70  & 43.02          & 72.86  & 54.09  & 43.41          & 33.75          & 75.82          & 47.60  & 39.56   \\
DeepSeek-VL2-Tiny-3B                                    & 21.95          & 27.96          & 24.59          & 13.69          & 46.52          & 16.97          & 24.87          & 19.89          & 29.83          & 21.12          & 24.73          & 16.79           \\
\rowcolor[rgb]{0.951,0.951,0.951} \textbf{F$^2$RVLM-Qwen2-VL-2B}   & 20.99          & 82.21          & 33.45          & 27.11          & 37.72          & 64.04          & 47.48          & 34.70          & 26.97          & 72.00          & 40.46          & 30.90           \\
\rowcolor[rgb]{0.951,0.951,0.951} \textbf{F$^2$RVLM-Qwen2.5-VL-3B} & 39.54          & 71.98          & 51.04 & 45.31 & 52.87  & 69.78          & 60.16 & 50.88 & 45.24  & 70.86          & 55.60 & 48.09  \\ 
\hline
LLaVA-1.5-7B-hf$^\ast$                                               & 18.16          & \textbf{89.80} & 30.21          & 9.78           & 24.66          & \textbf{74.21} & 37.02          & 36.20          & 20.91          & \textbf{81.27} & 33.61          & 22.99           \\
MiMo-7B-SFT                                                 & 40.06          & 52.03          & 45.27          & 37.60          & \underline{63.55}  & 42.16          & 50.69          & 43.82          & 49.14          & 46.58          & 47.98          & 40.71           \\
MiMo-7B-RL                                                  & 40.05          & 53.53          & 45.82          & 38.23          & \textbf{64.66} & 43.87          & 52.28          & 45.46          & 49.46          & 48.22          & 49.05          & 41.84           \\
Qwen2-VL-7B                                                   & 36.28          & 71.65          & 48.17          & 42.20          & 52.17          & 67.65          & 58.91          & 49.30          & 42.80          & 69.60          & 53.54          & 45.75           \\
Qwen2.5-VL-7B                                                 & 34.74          & 71.70          & 46.80          & 40.74          & 47.11          & 70.90  & 56.61          & 46.57          & 39.99          & 71.30          & 51.71          & 43.65           \\
InternVL3-8B$^\ast$                                                  & 31.05          & 80.78  & 44.86          & 25.98          & 37.76          & 70.90  & 49.28          & 47.33          & 34.08          & 75.52  & 47.07          & 36.65           \\
DeepSeek-VL2-Small-16B                                  & 40.13 & 24.92          & 30.75          & 24.93          & 57.98 & 16.00          & 25.07          & 24.09          & 47.43 & 19.48          & 27.91          & 24.51           \\
\rowcolor[rgb]{0.951,0.951,0.951} \textbf{F$^2$RVLM-Qwen2-VL-7B}   & \underline{52.36}  & 67.51          & \textbf{58.98} & \textbf{53.50} & 63.05          & 67.42          & \textbf{65.16} & \textbf{57.41} & \textbf{57.21} & 67.46          & \textbf{62.07} & \textbf{55.46}  \\
\rowcolor[rgb]{0.951,0.951,0.951} \textbf{F$^2$RVLM-Qwen2.5-VL-7B} & \textbf{52.44} & 64.51          & \underline{57.85}  & \underline{52.16}  & 57.46          & 65.82          & 61.36  & 52.63  & 54.83  & 65.16          & \underline{59.60}  & \underline{52.39}   \\
\bottomrule
\end{tabular}
}
\vspace{-3pt}  
\caption{Comparison with popular VLMs on the WeChat test set. We report Precision, Recall, F1, and MCC for utterance, image, and joint predictions. ``$\dagger$'' indicates zero-shot inference without MLDR fine-tuning. ``$\ast$'' indicates models limited by context length, evaluated via sliding-window inference.}
\label{tab:A3}
\end{table*}

\section{Appendix B: Implementation Details}
\subsection{B1. Training and Hyperparameter Settings}
We implement F$^2$RVLM using the ms-swift~\cite{zhao2024swiftascalablelightweightinfrastructure} framework, with Qwen-VL-Instruct as the backbone in 2B, 3B, and 7B configurations. Fine-tuning is conducted on the MLDR dataset, with 1,000 samples reserved for validation and the remaining for training. Among the training data, 25\% is used for SFT cold start, while the remaining is reserved for RFT. We adopt LoRA~\cite{hu2022lora} for parameter-efficient adaptation during both the SFT and RFT stages. During GRPO fine-tuning in RFT, we sample $G = 8$ candidate responses for each instance. The reward weights are set to $\lambda_{\text{utt}} = \lambda_{\text{img}} = 0.5$, and the exponential length penalty base is $\gamma = 0.95$. Curriculum sampling begins with 10\% of the easy and medium-level instances and gradually incorporates confusing and hard instances as training progresses. To enhance training stability and efficiency, we adopt the dynamic sampling strategy from DAPO~\cite{yu2025dapo}. Gradient checkpointing and FlashAttention are utilized to enhance training efficiency. All models are trained for 1 epoch on 8$\times$A100 GPUs.

\subsection{B2. Model Comparison Settings}
We evaluate F$^2$RVLM against a comprehensive set of VLMs in three categories: (1) \textbf{Proprietary models}, including Claude-Sonnet-4, GPT-4o, Gemini-2.5-Flash, and Doubao-Seed-1.6, are accessed via public APIs in inference-only mode without any fine-tuning. (2) \textbf{Open-source generation-based models}, including LLaVa, Qwen-VL-series, DeepSeek-VL2-series, InternVL3-series, Ovis2, Owl3, and MiMo-VL, are built from official checkpoints and fine-tuned on the MLDR dataset via SFT. All SFT models adopt the same prompt template as F$^2$RVLM and are trained in the ms-swift framework with LoRA for parameter-efficient adaptation. They share the same training configuration as F$^2$RVLM’s SFT stage, except that all training data are used for SFT and models are trained for one epoch. (3) \textbf{Embedding-based models}, including CLIP, BLIP-2, E5-V, and GME, are evaluated in inference-only mode. We compute the embedding of the query and each utterance/image in the dialogue, then apply a similarity threshold to retrieve relevant fragments, reporting the best results across all thresholds.

\subsection{B3. Sliding Window Inference for Long Contexts}
Some open-source generation-based models (e.g., Ovis2-2B, LLaVA-1.5-7B, and InternVL3-series) are constrained by limited context length and cannot directly process long-form dialogues in a single pass. To address this limitation, we adopt a sliding window inference strategy during evaluation. Specifically, the dialogue is segmented into overlapping windows of 35 turns, with an overlap of 15 turns between adjacent windows. Each window is independently processed by the model, and the fragment-level predictions are collected. For overlapping segments, we take the union of predicted results across windows to ensure comprehensive coverage and mitigate boundary truncation effects. This approach enables fair evaluation of models with limited context capacity while maintaining retrieval completeness.

\begin{table*}[t]
\centering
\resizebox{1.0\linewidth}{!}{
\begin{tabular}{r|cccc|cccc|cccc} 
\toprule
\multirow{2}{*}{Model}                                                  & \multicolumn{4}{c|}{Utterance Retrieval}                          & \multicolumn{4}{c|}{Image Retrieval}                              & \multicolumn{4}{c}{Joint Retrieval}                                \\ 
\cline{2-13}
& Precision      & Recall         & F1             & MCC            & Precision      & Recall         & F1             & MCC            & Precision      & Recall         & F1             & MCC             \\ 
\hline
Qwen2.5-VL-72B$^\dagger$                                                          & 30.31          & 32.04          & 31.15          & 21.57          & 34.23          & \underline{73.56}  & 46.72          & 36.60          & 32.15          & 44.64          & 38.94          & 29.08           \\
Claude-4$^\dagger$                                                                 & \underline{59.09}  & 39.46          & \underline{47.32}          & \underline{42.82}  & 59.05          & 59.62          & \underline{59.33}  & \underline{51.83}  & \textbf{59.07} & 47.49          & \underline{53.32}  & \underline{47.32}   \\
GPT-4o$^\dagger$                                                                   & 64.18          & 34.54          & 44.91          & 42.23          & 53.69          & 62.98          & 57.96          & 49.74          & \underline{58.47}  & 44.61          & 51.43          & 45.99           \\
Gemini-2.5$^\dagger$                                                               & \textbf{59.94} & 26.56          & 36.81          & 35.00          & 52.25          & 47.21          & 49.60          & 41.13          & 55.83          & 33.99          & 43.20          & 38.06           \\ 
\hline
mPLUG-Owl3-2B                                                                 & 22.32          & 16.92          & 19.25          & 10.14          & 16.71          & 29.81          & 21.42          & 5.13           & 19.11          & 21.59          & 20.33          & 7.64            \\
Qwen2-VL-2B                                                            & 15.90          & \textbf{86.34} & 26.86          & 16.69          & 27.57          & \textbf{81.25} & 41.17          & 30.58          & 20.17          & \textbf{83.72} & 34.01          & 23.64           \\
Qwen2.5-VL-3B                                                          & 23.78          & 76.42          & 36.27          & 28.87          & 34.47          & 72.60          & 46.75          & 36.50          & 28.15          & 74.46          & 41.51          & 32.68           \\
Mimo-7B-RL                                                           & 37.44          & 40.92          & 39.10          & 30.49          & \textbf{64.75} & 37.98          & 47.88          & 43.04          & 47.44          & 39.39          & 43.49          & 36.77           \\
Qwen2-VL-7B                                                            & 38.85          & 56.66          & 46.09          & 38.14          & 47.47          & 58.65          & 52.47          & 43.00          & 42.73          & 57.64          & 49.28          & 40.57           \\
Qwen2.5-VL-7B                                                          & 34.82          & 63.52          & 44.98          & 37.27          & 44.90          & 67.79          & 54.02          & 44.91          & 39.22          & 65.59          & 49.50          & 41.09           \\
DeepSeek-VL2-Small-16B                                                   & 26.43          & 14.66          & 18.86          & 11.97          & 55.88          & 9.69           & 16.52          & 18.72          & 35.89          & 11.67          & 17.69          & 15.35           \\
\rowcolor[rgb]{0.951,0.951,0.951} \textbf{\textbf{F$^2$RVLM-Qwen2-VL-2B}}   & 17.03          & \underline{86.27}  & 28.45          & 19.51          & 29.80          & 78.37          & 43.18          & 32.78          & 21.68          & \underline{82.13}  & 35.81          & 26.14           \\
\rowcolor[rgb]{0.951,0.951,0.951} \textbf{\textbf{F$^2$RVLM-Qwen2.5-VL-3B}} & 30.25          & 70.18          & 42.28          & 35.00          & 42.11          & 73.08          & 53.43          & 44.60          & 35.21          & 71.60          & 47.85          & 39.80           \\
\rowcolor[rgb]{0.951,0.951,0.951} \textbf{F$^2$RVLM-Qwen2-VL-7B}            & 50.19          & 53.81          & \textbf{51.94} & \textbf{45.23} & \underline{62.15}  & 63.94          & \textbf{63.03} & \textbf{56.15} & 55.54          & 58.44          & \textbf{57.49} & \textbf{50.69}  \\
\bottomrule
\end{tabular}
}
\vspace{-3pt}  
\caption{Evaluation results on the English-translated WeChat test set. We report precision, recall, F1, and MCC for utterance, image, and joint retrieval. ``$\dagger$'' indicates zero-shot inference without MLDR fine-tuning.}
\label{tab:A4}
\end{table*}

\begin{table}[t]
\centering
\resizebox{1.0\linewidth}{!}{
\begin{tabular}{c|cc|cc} 
\toprule
\multirow{2}{*}{Training Strategy} & \multicolumn{2}{c|}{In-domain} & \multicolumn{2}{c}{Real-domain}  \\ 
\cline{2-5}
& F1(\%)    & MCC(\%)                              & F1(\%)    & MCC(\%)                               \\ 
\hline
All SFT                            & 85.57 & 77.98                            & 47.60 & 39.56                             \\
All GRPO(0.1\% SFT)                & 82.14      & 73.29                                 & 45.47      & 37.81                                   \\
\hline
5\%SFT+95\%GRPO                    & 84.58      & 78.65                                & 53.64      & 42.83                                   \\
10\%SFT+90\%GRPO                   & 86.44       & 79.26                                 & 54.51      &  47.18                                 \\
25\%SFT+75\%GRPO                   & \underline{87.00} & \underline{80.19}                            & \textbf{55.60} & \textbf{48.09}                             \\
50\%SFT+50\%GRPO                   & \textbf{87.37}      & \textbf{81.05}                                  & \underline{54.32}      & \underline{47.33}                                  \\
\bottomrule
\end{tabular}
}
\vspace{-3pt}  
\caption{Ablation study of different SFT and GRPO data ratios in F$^2$RVLM on MLDR validation and WeChat test sets.} \label{tab:A5}
\end{table}

\section{Appendix C: Experimental Results and Analysis}
\subsection{C1. Detailed Comparison of Quantitative Results}
Due to space limitations in the main text, Table 2 only presents the overall performance of selected models on the MLDR and WeChat datasets in terms of joint precision, recall, F1, and MCC. To provide a more comprehensive evaluation, this section reports the full set of metrics, including separate results for utterance-level, image-level, and joint fragment-level retrieval across all models. Each category includes precision, recall, F1, and MCC, offering a finer-grained comparison of model capabilities.

\noindent \textbf{MLDR Validation Set.} Table~\ref{tab:A2} presents the comparison results on the MLDR validation set, embedding-based models (e.g., CLIP, BLIP-2, E5-V, GME) generally underperform across all metrics. Without task-specific supervision, they struggle to align utterance and image semantics, resulting in imbalanced retrieval (e.g., BLIP-2 shows extremely high utterance recall but almost zero image recall and MCC). Proprietary models (e.g., Claude, GPT-4o, Gemini, Doubao) exhibit moderate performance, achieving reasonable F1 and MCC scores in joint retrieval, but their recall remains limited due to the lack of task adaptation. In contrast, open-source models fine-tuned with SFT on MLDR (e.g., Qwen2.5-VL, InternVL3, Owl3) demonstrate substantial improvements, balancing precision and recall and achieving higher MCC scores. Our proposed F$^2$RVLM series outperforms all baselines in nearly all metrics. Notably, F$^2$RVLM-7B achieves the best joint F1 (87.25\%) and MCC (80.60\%), and maintains strong utterance and image-level performance. Even the lighter-weight 2B and 3B variants surpass larger models such as MiMo-7B and InternVL3-8B, highlighting the effectiveness of GRPO-based fine-tuning and difficulty-aware curriculum sampling in improving fine-grained retrieval accuracy.

\noindent \textbf{WeChat Test Set.} Table~\ref{tab:A3} further evaluates the generalization performance of all models on the WeChat test set, which features longer contexts, more linguistic noise, and broader domain shifts, posing a more challenging evaluation scenario. Results show that fine-tuning on MLDR significantly improves cross-domain fragment retrieval. For instance, Qwen2.5-VL-7B improves its joint F1 from 12.58\% (MCC = 0.00\%) in zero-shot to 51.71\% (MCC = 43.65\%) after SFT on MLDR, over a fourfold increase. MiMo-7B also benefits notably from task-specific fine-tuning. These findings confirm that MLDR supervision enhances not only in-domain performance but also out-of-domain alignment and reasoning. In contrast, proprietary models such as GPT-4o and Gemini-2.5, without access to MLDR fine-tuning, still exhibit strong performance due to their massive scale and diverse pretraining corpora. Nevertheless, F$^2$RVLM maintains consistently strong results across all model sizes. The 7B variant achieves the best joint F1 (62.07\%) and MCC (55.46\%), while even the lightweight 2B and 3B variants outperform larger baselines. This underscores the effectiveness of our GRPO-based optimization and curriculum-guided training in enhancing fine-grained semantic alignment and context reasoning under domain shifts.

\subsection{C2. Comparison of Qualitative Results}
We provide qualitative comparisons on the MLDR validation set and the WeChat test set to illustrate fragment-level retrieval performance.

\begin{figure*}[!ht]
\centering
\includegraphics[width=0.85\linewidth]{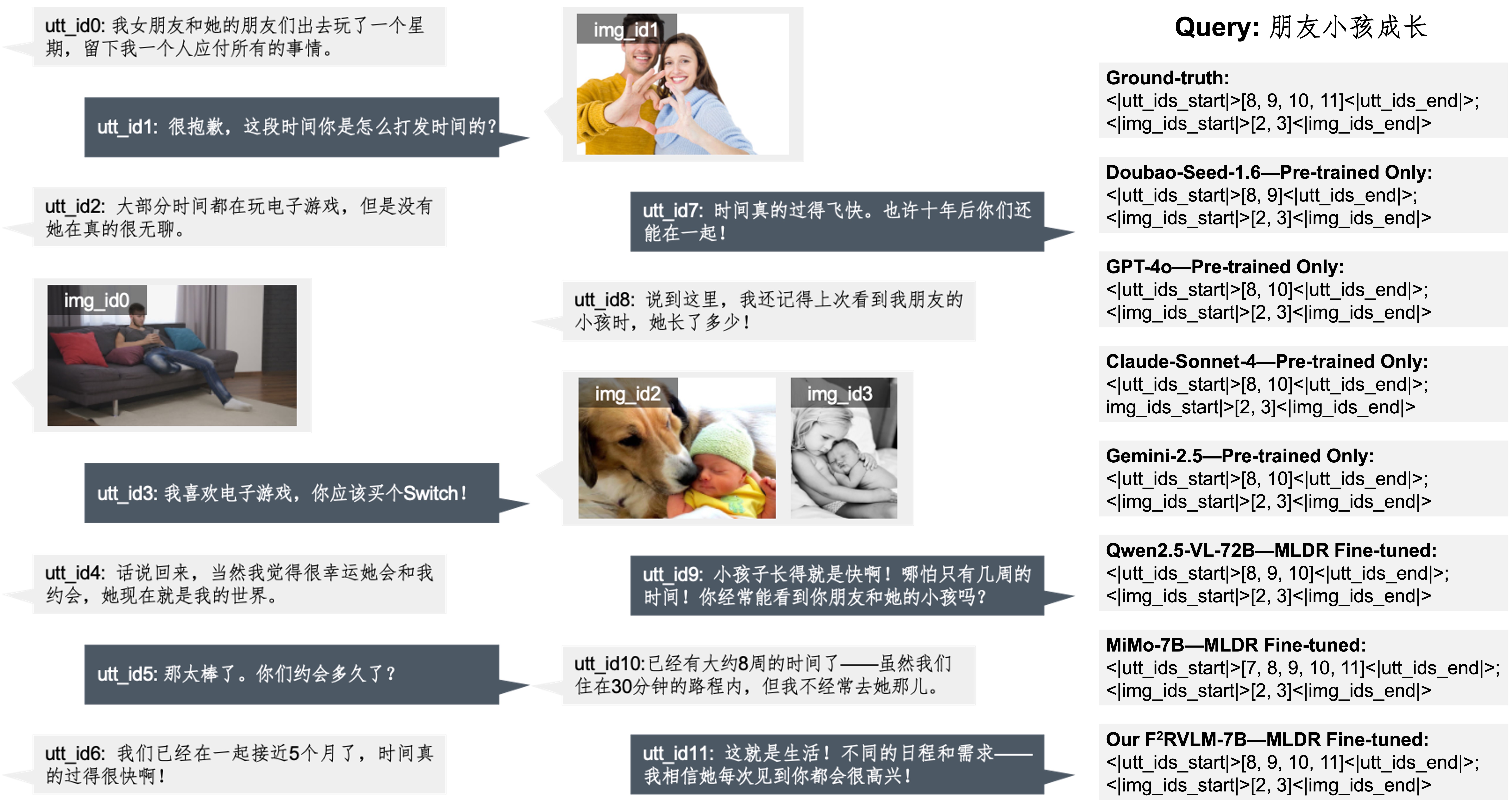}
\caption{Qualitative comparison on the MLDR validation set. Given a user query, we visualize one representative case of retrieved fragments from various models. Pre-trained models often retrieve semantically relevant but incomplete or disjointed fragments. MLDR fine-tuned models improve alignment but may still miss contextual boundaries. Our F$^2$RVLM-7B achieves the most coherent and complete retrieval, accurately aligning both utterances and images with the intended semantics.}
\label{fig:A4}
\end{figure*}

\begin{figure*}[!ht]
\centering
\includegraphics[width=1.0\linewidth]{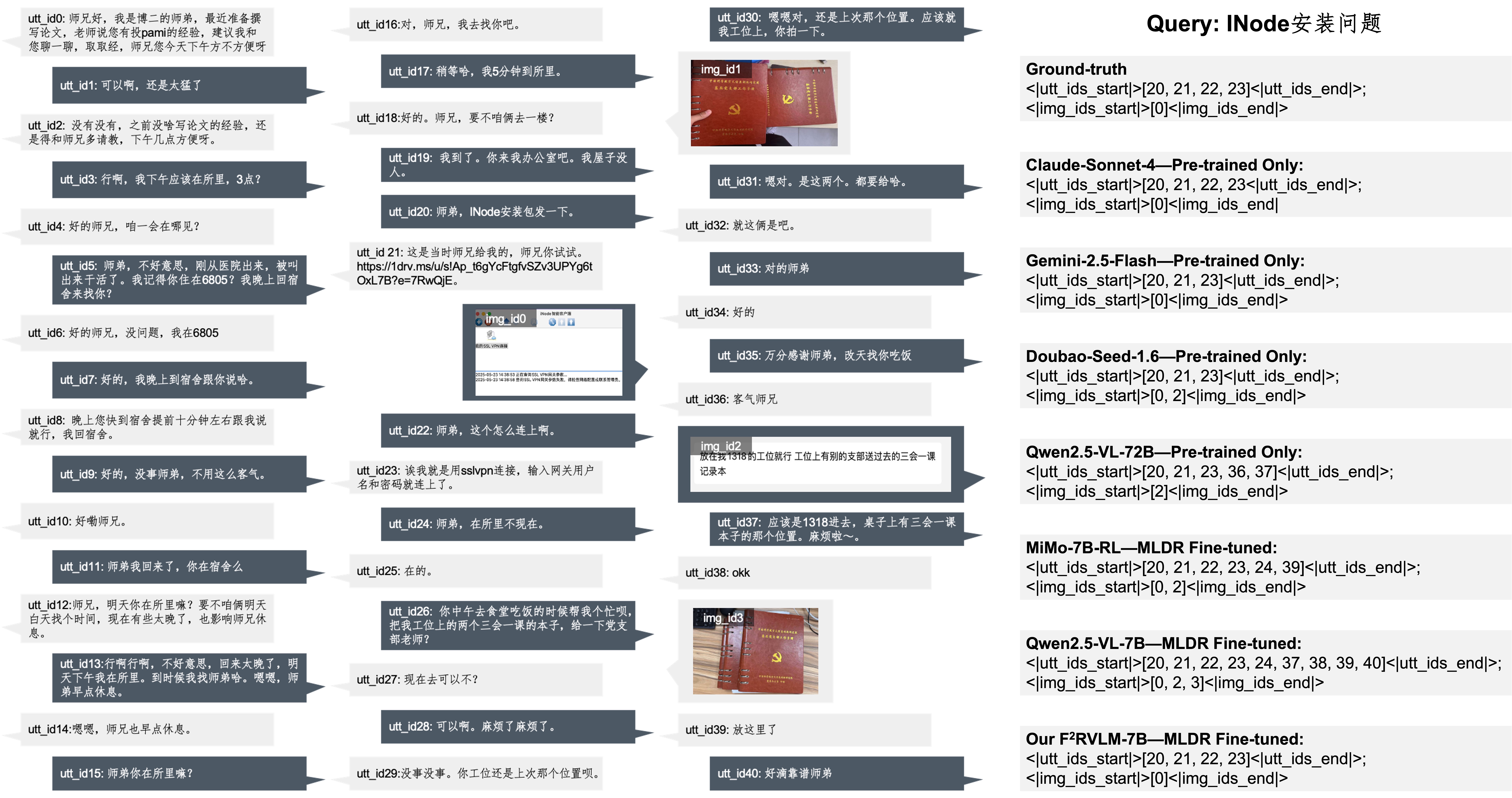}
\vspace{-15pt}  
\caption{Case 1 of qualitative comparison on the WeChat test set, focusing on a personal-life dialogue scenario.}
\label{fig:A5}
\end{figure*}

\begin{figure*}[!ht]
\centering
\includegraphics[width=1.0\linewidth]{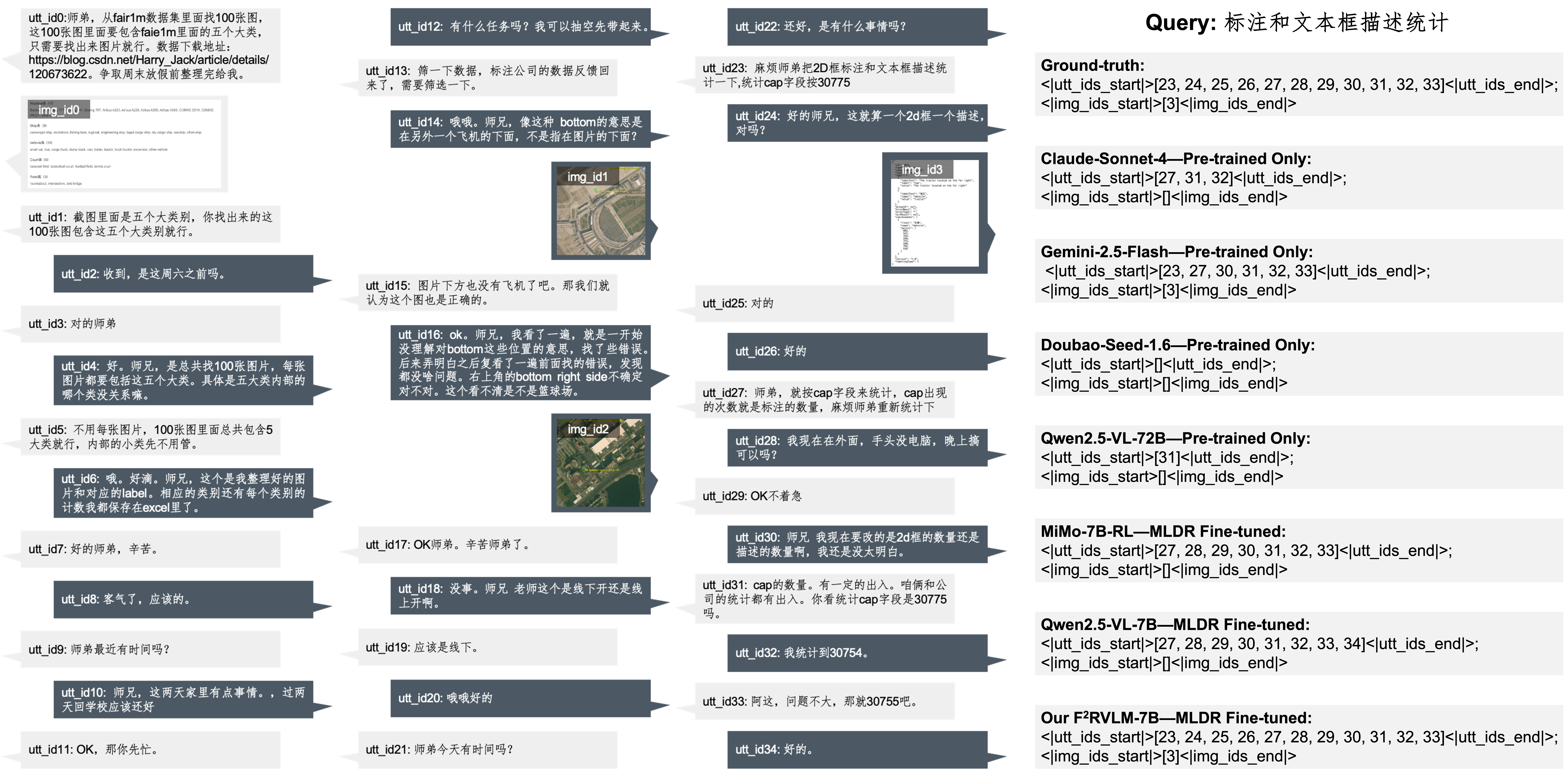}
\vspace{-15pt}  
\caption{Case 2 of qualitative comparison on the WeChat test set, illustrating fragment retrieval in a technical-support dialogue scenario.}
\label{fig:A6}
\end{figure*}

On the MLDR validation set (Fig.\ref{fig:A4}), our F$^2$RVLM accurately retrieves semantically complete and coherent fragments that align well with human annotations. It outperforms closed-source or open-source large models that often miss key utterances or include off-topic content. On the WeChat test set (Fig.\ref{fig:A5}–\ref{fig:A6}), our model demonstrates better robustness in informal, multi-topic dialogues. While closed-source and open-source models tend to produce partial or noisy results, F$^2$RVLM captures more complete, relevant, and coherent fragments, effectively combining verbal and visual cues even in complex real-world cases. These results further validate the model’s superior capability in real-world fragment-level retrieval.

\subsection{C3. Comparison of Cross-Lingual Generalization }
To assess the cross-lingual robustness, we evaluate various VLMs on an English-translated version of the WeChat test set. Specifically, we randomly sample 200 dialogues from the original set and translate each utterance into English using DeepL Translator to preserve semantic fidelity. The translated set maintains the original structure and multimodal content, enabling fair comparison across languages. 

As depicted in Table~\ref{tab:A4}, even though models like Claude-4 and GPT-4o exhibit strong zero-shot performance due to extensive pre-training on English corpora, our F$^2$RVLM-7B achieves the highest overall joint F1 (57.49\%) and MCC (50.69\%). Even the 3B variant of F$^2$RVLM outperforms larger models like MiMo-7B and Qwen2.5-VL-7B, confirming its strong generalization with fewer parameters. Importantly, unlike some VLMs that exhibit skewed retrieval behavior (e.g., high precision but low recall, or vice versa), F$^2$RVLM achieves a more balanced trade-off between precision and recall. This balance results in consistently higher F1 and MCC scores, highlighting the model’s ability to perform fine-grained fragment retrieval across languages. The above results suggest that the reward-guided fine-tuning strategy in F$^2$RVLM effectively transfers to different linguistic settings, supporting fine-grained fragment alignment and reasoning beyond Chinese-language contexts.

\subsection{C4. Human Subjective Evaluation Criteria}
To assess the human-perceived quality of fragment-level retrieval, we randomly sample 200 dialogues and ask expert annotators to compare model outputs under identical input contexts. Each annotator is instructed to select the best-performing model according to the following three criteria:
\begin{itemize}
    \item \textbf{Coverage,} which measures whether the retrieved fragment fully captures all key information relevant to the user query. It emphasizes completeness, particularly the inclusion of essential contextual elements. Fragments that contain more content can still receive high coverage scores as long as they thoroughly address the query without omitting important details.
    \item \textbf{Relevance,} which evaluates how well the retrieved content semantically aligns with the user query. Every utterance and image in the fragment should directly relate to the query’s core intent. Even if some parts are on-topic, the presence of off-topic or irrelevant content will reduce the relevance score.
    \item \textbf{Coherence,} which assesses the logical flow and clarity of the retrieved fragment itself. It focuses on whether the sentences and images are naturally connected, well-organized, and easy to understand when read as a self-contained unit. This evaluation does not consider the broader dialogue context, but strictly the internal consistency of the retrieved segment.
\end{itemize}
If multiple models produce identical and top-performing outputs for a given dialogue, annotators will select all such models. As a result, the total number of top-choice selections may exceed the number of dialogues evaluated.

\subsection{C5. Ablation Studies on SFT and GRPO Ratio}
Table~\ref{tab:A5} explores the impact of different proportions of SFT and GRPO-based RFT on model performance. We observe the following: (1) Using GRPO alone underperforms compared to using only SFT, highlighting the indispensable role of supervised learning for cold-start in providing stable and accurate training signals; (2) As the proportion of GRPO increases, overall performance gradually surpasses that of pure SFT, demonstrating the effectiveness of RFT for dialogue fragment retrieval; (3) A 25\% SFT and 75\% GRPO ratio achieves the best results on the real-domain test set, while increasing SFT to 50\% slightly improves in-domain validation performance but compromises generalization to real-world scenarios.
\end{document}